\documentclass[lettersize,journal]{IEEEtran}
\usepackage{amsmath,amsfonts}
\usepackage{algorithmic}
\usepackage{algorithm}
\usepackage{array}
\usepackage[caption=false,font=normalsize,labelfont=sf,textfont=sf]{subfig}
\usepackage{textcomp}
\usepackage{stfloats}
\usepackage{url}
\usepackage{verbatim}
\usepackage{graphicx}
\usepackage{cite}
\usepackage{algorithm,algorithmic}
\usepackage{multirow,booktabs,pifont}
\usepackage{amsmath}
\usepackage{colortbl}
\usepackage{xcolor}
\usepackage{amssymb}
\usepackage{marvosym}
\usepackage{amsmath}

\usepackage{float}
\usepackage{placeins} % 提供 \FloatBarrier

\definecolor{lightgray}{gray}{0.9}
\definecolor{lightblue}{rgb}{0.9, 0.9, 1.0}

\begin{document}
\title{MEGA: Second-Order Gradient Alignment for Catastrophic Forgetting Mitigation in GFSCIL}

\author{Jinhui Pang, Changqing Lin, Hao Lin, Zhihui Zhang, Weiping Ding, Yu Liu, Xiaoshuai Hao

\thanks{Jinhui Pang, Changqing Lin, Hao Lin and Zhihui Zhang are with the Beijing Institute of Technology, Beijing, China (e-mail: pangjinhui@bit.edu.cn, lcq2000@bit.edu.cn, 1120222396@bit.edu.cn,3220231441@bit.edu.cn).
}

\thanks{Yu Liu is with the Department of Biomedical Engineering, Hefei University of Technology,  Hefei, China (e-mail: yuliu@hfut.edu.cn).}

\thanks{Weiping Ding is the School of Artificial Intelligence and Computer Science, Nantong University, Nantong, 226019, China, and also the Faculty of Data Science, City University of Macau, Macau 999078, China (e-mail: dwp9988@163.com).}

\thanks{Xiaoshuai Hao is with the Beijing Academy of Artificial Intelligence, Beijing, China (e-mail: xshao@baai.ac.cn).}

}

\markboth{Journal of \LaTeX\ Class Files,~Vol.~14, No.~8, August~2021}%
{Shell \MakeLowercase{\textit{et al.}}: A Sample Article Using IEEEtran.cls for IEEE Journals}

\IEEEpubid{0000--0000/00\$00.00~\copyright~2021 IEEE}

\maketitle

\begin{abstract}
Graph Few-Shot Class-Incremental Learning (GFSCIL) enables models to continually learn from limited samples of novel tasks after initial training on a large base dataset. 
Existing GFSCIL approaches typically utilize Prototypical Networks (PNs) for metric-based class representations and fine-tune the model during the incremental learning stage.
However, these PN-based methods oversimplify learning via novel query set fine-tuning and fail to integrate Graph Continual Learning (GCL) techniques due to architectural constraints.
To address these challenges, we propose a more rigorous and practical setting for GFSCIL that excludes query sets during the incremental training phase. 
Building on this foundation, we introduce Model-Agnostic Meta Graph Continual Learning (\textit{MEGA}), aimed at effectively alleviating catastrophic forgetting for GFSCIL.
Specifically, by calculating the incremental second-order gradient during the meta-training stage, we endow the model to learn high-quality priors that enhance incremental learning by aligning its behaviors across both the meta-training and incremental learning stages.
Extensive experiments on four mainstream graph datasets demonstrate that MEGA achieves state-of-the-art results and enhances the effectiveness of various GCL methods in GFSCIL. 
We believe that our proposed MEGA serves as a model-agnostic GFSCIL paradigm, paving the way for future research.
\end{abstract}


\section{Introduction}

\IEEEPARstart{G}{raph} Neural Networks (GNNs) \cite{GNN,GCN,GraphSAGE} have emerged as an innovative approach in machine learning, showing significant promise in various fields, including social network analysis \cite{FakeNews}, recommendation systems \cite{recommendation}, bioinformatics \cite{bioinfo}, and molecular structure prediction \cite{Protein}.
However, the inherent complexity, sparsity, and dynamic nature of graph data pose unique challenges, such as acquiring labeled data, adapting to new categories, and managing computational costs.
To address these challenges, researchers are exploring few-shot learning \cite{Meta-gnn,GPN,GraphMetaLearning,MI-GNN} and class-incremental learning \cite{CGL2,RieGrace,Cat,PI-GNN} in the graph domain.
Few-shot learning aims to minimize the reliance on labeled data and improve model generalization, whereas class-incremental learning enables adaptation to evolving graph structures and supports the continuous learning of new node categories.
Therefore, Graph Few-Shot Class-Incremental Learning (GFSCIL) has emerged, integrating these approaches to address graph-specific challenges and enhance the practicality and adaptability of GNNs \cite{HAG-Meta,Geometer}.
Different from classic Few-Shot Class-Incremental Learning (FSCIL) scenarios, GFSCIL requires the model to maintain its ability to mitigate catastrophic forgetting, even as the graph structure evolves dynamically.
This is rather challenging and thus becomes the motivation of our work.

GFSCIL requires models to  maintain continual learning capabilities across a series of novel tasks with limited samples, following initial training on a large dataset of base classes.
This task faces two primary challenges: first, the scarcity of examples for newly introduced categories can lead to model overfitting; second, the inability to access graph data from previously learned categories during incremental stages results in catastrophic forgetting of established class knowledge.
Existing works \cite{HAG-Meta,Geometer} typically employ Prototypical Networks (PNs) \cite{PNs} to address the overfitting issue while mitigating catastrophic forgetting through model fine-tuning during incremental learning stages.
In standard FSCIL settings, models can only use the query set from the base dataset for training, and must use the incremental stage's query set solely for testing. 
This fundamentally differs from current methods: although existing approaches achieve good performance on public graph datasets by using novel query sets during incremental training, this practice makes GFSCIL tasks easier than real-world applications require, reducing practical usefulness.
Moreover, PN-based models, which rely on specific architectural frameworks to address the issue, lack flexibility and do not effectively utilize established GCL \cite{CGLB,SSM, LwF} methods in GFSCIL scenarios.

\IEEEpubidadjcol

\begin{figure}[t]
\centering
\includegraphics[width=0.99\columnwidth]{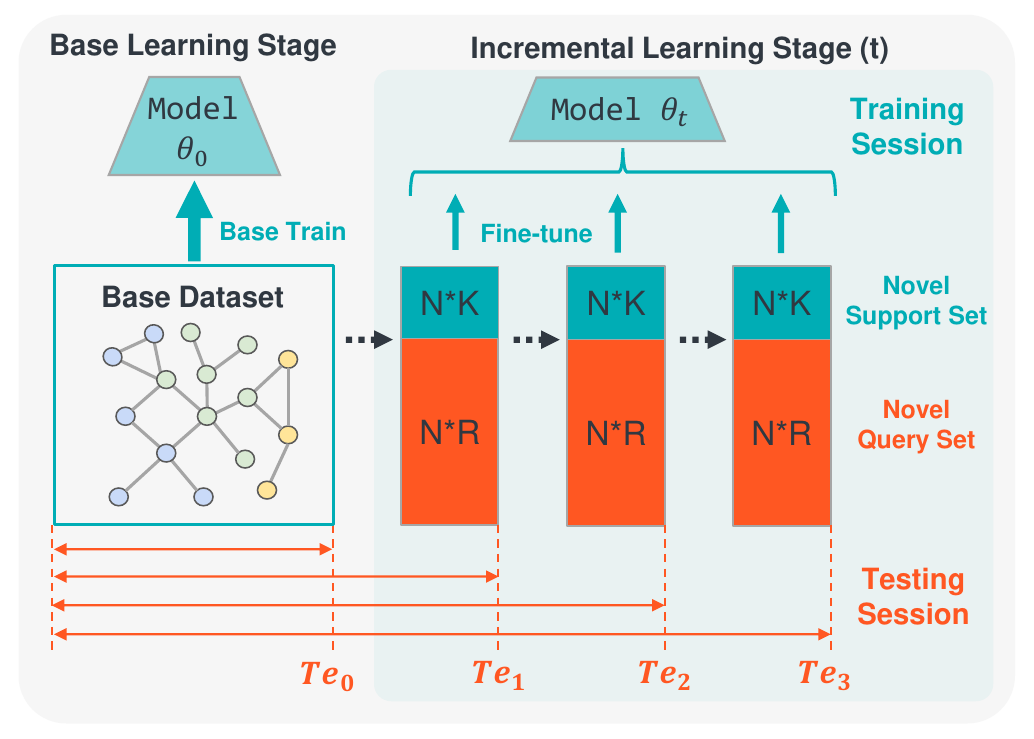}
%%\vspace{-1em}
\caption{\textbf{Graph few-shot class-incremental learning problem ($N$-way $K$-shot $R$-query). }
In our proposed setting, each novel task is divided into a support set and a query set, both containing $N$ classes. 
Each class within these sets comprises $K$ sample nodes in the support set and $R$ sample nodes in the query set, respectively.
}
\label{fig:setting}
%\vspace{-1em}
\end{figure}

To address the aforementioned issues, we first introduce a more rigorous setting for GFSCIL that restricts the use of novel query sets during incremental training, effectively addressing  the query set leakage problem. 
As shown in Figure \ref{fig:setting}, our proposed setting divides each novel task into a support set and a query set, where the model is trained exclusively on the novel support sets and evaluated on the novel query sets.
Importantly, since we do not modify the settings of the base learning stage, the model still acquires highly generalizable initialization parameters through extensive training with large-scale data during base training.
Building on this foundation, we propose a novel approach called Model-Agnostic Meta Graph Continual Learning (\textbf{\textit{MEGA}}), which divides the model training process into a meta-training stage and an incremental learning stage, while also maintaining behavioral consistency across these stages.
Specifically, in the meta-training stage, we introduce an incremental second-order gradient approach to identify optimal initial parameters for continual learning, adapting second-order gradient descent to accommodate an incremental task sequence.
Moreover, during the incremental training stage, we utilize the same loss functions as in the meta-training stage to align model behaviors across both stages, ensuring efficient use of high-quality priors for consistent continual learning.
To demonstrate MEGA's ability to mitigate catastrophic forgetting for graph data, we conduct extensive experiments on four prominent graph datasets, validating its effectiveness in extracting high-quality priors from graph data.
Importantly, our proposed MEGA establishes a model-agnostic GFSCIL paradigm, offering simple yet effective plug-and-play techniques compatible with existing GCL methods, paving the way for future research.

In summary, our contributions are as follows:
\begin{itemize}
    \item To address query set leakage in existing GFSCIL settings, we propose a more rigorous setting where the model trains exclusively on the novel support sets during incremental training, thereby elevating task complexity and better aligning with real-world applicability.

    \item To effectively alleviate catastrophic forgetting in our proposed challenging GFSCIL setting, we introduce MEGA, a model-agnostic and versatile framework that leverages incremental second-order gradient computation to derive high-quality priors, thereby mitigating overfitting and enhancing generalization.
    
    \item Compared with the state-of-the-art methods, MEGA shows superior performance, achieving an improvement of 2.60\% to 7.43\% in the first novel task across four public graph datasets, demonstrating the superiority of our method.
\end{itemize}

\section{Related Work}
\noindent\textbf{Few-shot Class-incremental Learning.} 
Few-shot Class-Incremental Learning (FSCIL) presents a significant challenge for models, requiring them to learn new tasks from minimal labeled data while retaining knowledge of previously mastered tasks \cite{IDLVQ,FACT,FSIL-GAN,SoftNet}. 
This area has been extensively studied in computer vision \cite{ServeyFSCIL}. 
The first framework to address the FSCIL challenge, known as TOPIC, is introduced by \cite{TOPIC}, which utilizes a neural gas network to learn the topological structure of the feature space.
Following this, \cite{S3C} proposes the Self-Supervised Stochastic Classifier (S3C) to mitigate overfitting on novel classes.
In the realm of graphs, HAG-Meta \cite{HAG-Meta} and GEOMETER \cite{Geometer} have pioneered the extension of FSCIL to graph few-shot class-incremental learning (GFSCIL). 
Both approaches leverage prototypical networks (PNs) \cite{PNs,GPN} to tackle the challenge of limited sample sizes and adapt to incremental learning scenarios through fine-tuning on the query sets of novel classes.
However, existing GFSCIL methods primarily focus on addressing the problem through specific model architectures, such as PNs, with limited exploration of adapting well-established GCL methods to GFSCIL scenarios. 
In contrast, the proposed MEGA adopts a model-agnostic design, enabling seamless integration of state-of-the-art GCL methods by leveraging their loss functions at each training stage, thereby facilitating continuous advancements in GFSCIL performance.

\noindent\textbf{Graph Meta Learning.}
Graph meta-learning is a widely used approach for tackling the Graph Few-Shot Learning (GFSL) problem \cite{SurveyGFSL,SurveyMetaL, Meta-GPS++}. 
It employs various meta-learning \cite{RALE, GFL} techniques to derive high-quality priors from graph data \cite{GraphMetaLearning, FAAN}.
Meta-GNN \cite{Meta-gnn} is the first to incorporate the MAML \cite{MAML} algorithm into the graph domain to address the GFSL problem.
G-Meta \cite{g-meta} proposes using local subgraphs to learn knowledge representations that can be transferred across tasks.
SELAR \cite{SELAR} further highlights the effectiveness of GNNs through a self-supervised auxiliary learning framework integrated with meta-learning.
However, these methods primarily focus on the GFSL problem and often overlook the need for continual training in real-world scenarios.
Consequently, they experience catastrophic forgetting when applied to GFSCIL. 
In contrast to previous approaches, we introduce optimization-based meta-learning methods to GFSCIL scenarios, allowing the model to learn priors while effectively mitigating catastrophic forgetting.

\section{Method}

\subsection{Preliminaries}

\noindent\textbf{Notations.}
Formally, for a graph $\mathcal{G=(V,E)}$, where $\mathcal{V}$ and $\mathcal{E}$ denote the set of nodes and edges, we study the node classification problem. Each node $v\in \mathcal{V}$ has a feature vector $x(v)$ and a category label $y(v) \in \mathcal{Y}$, where $\mathcal{Y}=\{y^i\}_{i=1}^c$ is the label set and $c$ is the number of classes.
The GFSCIL setting is characterized by a sequence of disjoint tasks, denoted as $\{\mathcal{T}_i\}_{i=0}^n$, where $n$ represents the number of novel tasks. In this sequence, $\mathcal{T}_0$ is designated as the base task, while $\{\mathcal{T}_i\}_{i=1}^n$ represents the set of novel tasks.
Each task $\mathcal{T}_i=(\mathcal{G}_i,\mathcal{Y}_i)$ comprises a label space including multiple non-overlapping classes  $\mathcal{Y}_i=\{y^j\}_{j=1}^{c_i}$ and $c_i$ is the number of classes in task $\mathcal{T}_i$. 
We define the node set of $\mathcal{T}_i$ as $\mathcal{V}_i=\{v|y(v)\in\mathcal{Y}_i,v\in\mathcal{V}\}$. 
Furthermore, we define $x(\mathcal{T}_i)$ and $y(\mathcal{T}_i)$ as the feature matrix and label vector for task $i$, respectively. Each row in $x(\mathcal{T}_i)$ represents the feature vector of a node in $\mathcal{V}_i$, while each element in $y(\mathcal{T}_i)$ corresponds to the label of a node in the same set.

\noindent\textbf{Problem Definition.} 
In this paper, we focus on the few-shot class-incremental node classification problem. For each task $\mathcal{T}_i$, we have a support set $\mathcal{S}_i\subset\mathcal{T}_i$ and a query set $\mathcal{Q}_i\subset\mathcal{T}_i$. 
The label spaces of $\mathcal{S}_i$ and $\mathcal{Q}_i$ are identical to that of $\mathcal{T}_i$, but the intersection of the node sets of $\mathcal{S}_i$ and $\mathcal{Q}_i$ is empty. 
There are two learning stages in the GFSCIL setting: the base learning stage and the incremental learning stage. 
In the base learning stage, the model is trained on $\mathcal{T}_0$ and tested on $\mathcal{Q}_0$. 
The number of nodes in $\mathcal{S}_0$ is relatively larger than other tasks, while the number of nodes in $\mathcal{Q}_0$ is identical to other tasks.
In the incremental learning stage, after training on $\mathcal{T}_0$, the model is continually trained on $\{\mathcal{S}\}_{i=1}^n$. 
Following training in $\mathcal{S}_i\ (i>0)$, we use $\bigcup_{j=0}^{i} \mathcal{Q}_j$ to evaluate its performance in both current and past tasks. 
For novel tasks $\mathcal{T}_i\ (0<i\leq n)$, when we denote $c_i$ as $N$, the number of nodes in each class in $\mathcal{S}_i$ as $K$ and the number of nodes in each class in $\mathcal{Q}_i$ as $R$, this setting is named as an \textbf{$N$-way, $K$-shot, $R$-query} GFSCIL setting. 
The goal of GFSCIL is to learn high-quality priors from the base task and maintain consistent high performance on both the base task and novel tasks. 

\begin{figure}[t]
\centering
\includegraphics[width=0.99\columnwidth]{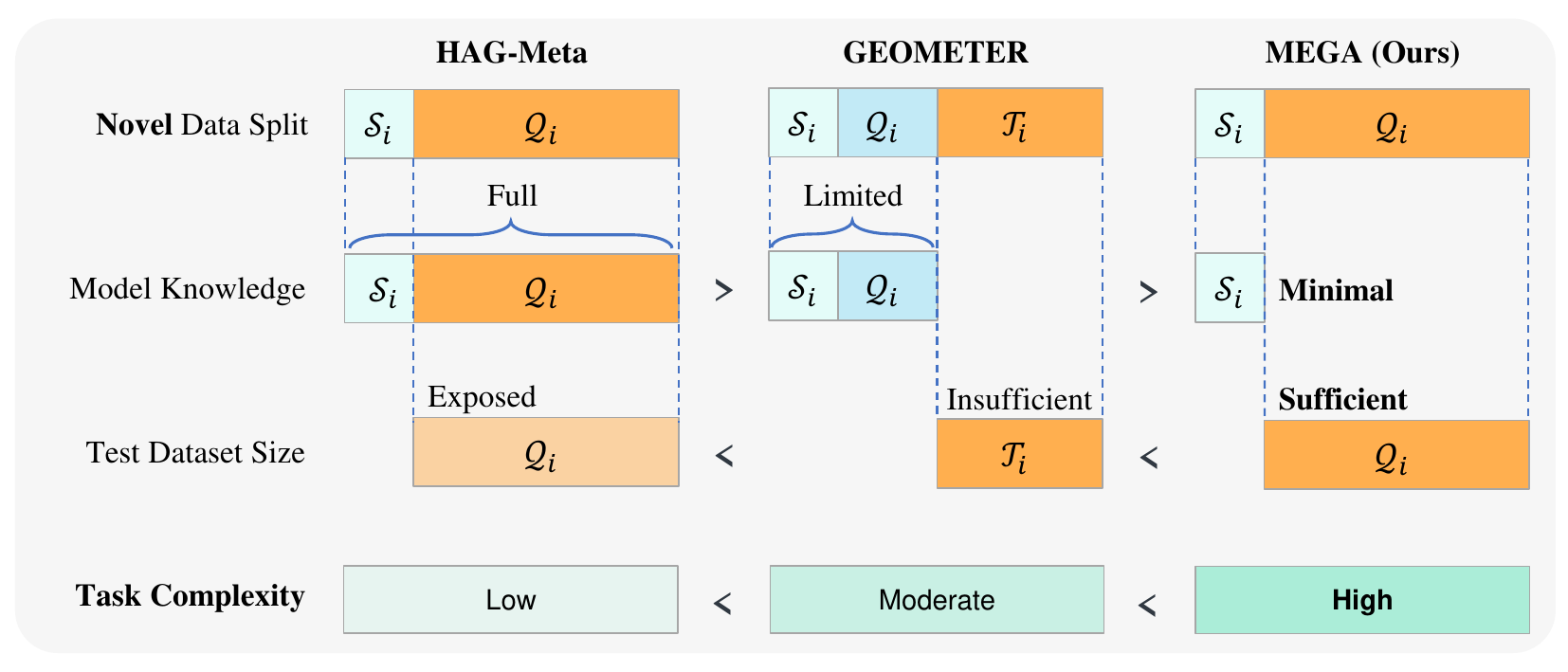} 
%%\vspace{-1em}
\caption{
Comparison of dataset partitioning strategies and task complexity in different GFSCIL settings.
$\mathcal{S}_i$, $\mathcal{Q}_i$, and $\mathcal{T}_i$ represent the support set, query set, and test set of the $i$-th novel task ($i > 0$), respectively.
Our proposed setting uses minimal model knowledge to infer as much unknown information as possible, resulting in the highest task complexity.
Notably, the distinction between the three settings lies specifically in the split of dataset during the incremental learning stage, while the base learning stage remains consistent. 
}
\label{fig:taskDifficulty}
%\vspace{-1em}
\end{figure}

\subsection{Task Complexity of GFSCIL}
Existing methods of GFSCIL often incorporate the query set of novel tasks during the incremental fine-tuning process, inadvertently reducing task complexity and straying from real-world scenarios.
As illustrated in Figure \ref{fig:taskDifficulty}, following the Meta-GNN configuration \cite{Meta-gnn}, HAG-Meta partitions each novel dataset into two subsets: the support set and the query set. Under normal circumstances, fine-tuning should only occur during the base learning stage. However, during the incremental learning stage, this approach still constructs class prototypes using the support set, subsequently fine-tunes the model on the query set, and ultimately evaluates the model's classification capability on the query set. Since the query set is already incorporated into the model’s knowledge, this task presents the lowest level of complexity.

Furthermore, GEOMETER divides each novel dataset into three subsets: the support set, the query set, and the test set. This method establishes class prototypes based on the support set, fine-tunes the model on the query set, and finally evaluates the model’s classification performance on the test set. Although the training and test sets are separated, the task remains easier than classic FSCIL due to the excessively large size of the training set.

Both previously proposed settings not only incorporate the query set during base training but also retain it in the incremental training, thereby lowering task complexity in graph-based FSCIL while inducing query set leakage.
To align the task complexity of GFSCIL with classic FSCIL tasks, we propose a more rigorous setting where novel datasets are divided into support and query sets, with the model trained exclusively on the novel support sets.
By leveraging minimal model knowledge to maximize the inference of unknown information, our proposed setting generates highly challenging tasks that more accurately evaluate GFSCIL methods in real-world few-shot scenarios, thereby facilitating the development of more robust and effective GFSCIL approaches for practical applications.

\begin{figure*}[t]
\centering
\includegraphics[width=0.99\textwidth]{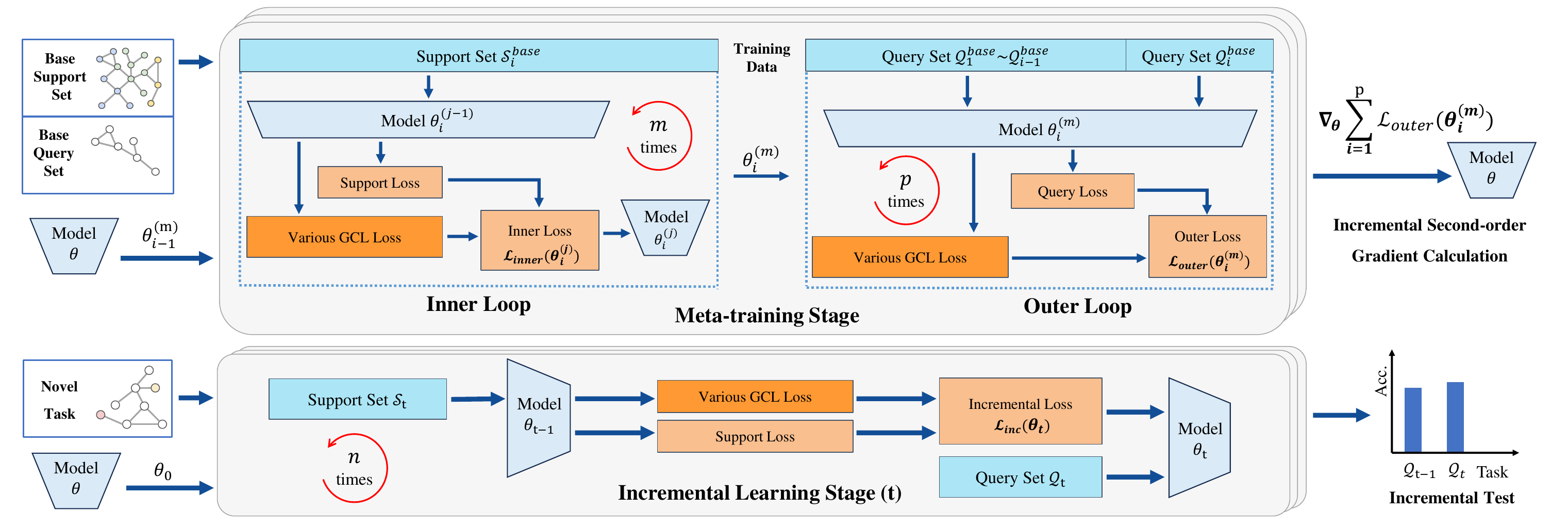} % Reduce the figure size so that it is slightly narrower than the column.
%%\vspace{-1.0em}
\caption{\textbf{Overall architecture of MEGA.} In the meta-training stage, we train a model by calculating the incremental second-order gradient. Then, in the incremental learning stage, the model is fine-tuned using the novel support set. By aligning the model's behavior across both stages, we enable it to acquire prior knowledge about optimal initial parameters across a task stream.}
\label{fig:framework}
%%\vspace{-1.0em}
\end{figure*}

\subsection{Model-Agnostic Meta Graph Continual Learning}
Most existing GCL methods mitigate catastrophic forgetting by modifying their loss functions. Similarly, current GFSCIL approaches achieve this goal through specifically designed loss functions, which places them on the same operational level as conventional GCL methods, thereby limiting their mutual compatibility. Furthermore, as Prototypical Networks (PNs) based approaches, their loss functions are inherently coupled with the PN architecture, making it challenging to incorporate other general GCL methodologies.

We address this limitation with Model-Agnostic Meta Graph Continual Learning (MEGA), which is inspired by MAML \cite{MAML}.
Figure \ref{fig:framework} presents the overall framework of our proposed MEGA method.
As a versatile and well-established meta-learning algorithm, MAML is typically used to identify optimal initial parameters for models on unseen tasks.
However, in GFSCIL, the model requires optimal initial parameters not for a single task, but across an entire stream of tasks. 
Therefore, simply applying MAML to GFSCIL does not effectively alleviate catastrophic forgetting.
This limitation stems from MAML's single-task optimization nature, which fails to capture inter-task correlations in continual learning.

To equip meta-learning algorithms with continual learning capabilities, we observe gradient descent similarities in MAML's inner/outer loops during meta-training and meta-testing. 
These three processes optimize the model using loss functions of the same structure, differing only in the input data.
Based on this observation, we hypothesize that a key aspect of MAML's ability to learn priors is ensuring consistent model behavior between the meta-training and meta-testing stages. 
This consistency activates the priors learned during meta-training in the meta-testing stage. 
Building on this hypothesis, we propose the MEGA method, driven by its core component, the Meta Continual Training Framework (\textbf{\textit{MCTF}}) — MEGA's meta-training stage learning algorithm. 
MCTF aligns learning behaviors across meta-training and incremental stages through incremental second-order gradient computation, which we now proceed to formally define.

\noindent\textbf{Incremental second-order gradient.} 
We first define an epoch of MCTF's meta-training stage as a double-loop. 
In the outer loop, at the beginning of each iteration, we first randomly sample an ordered task sequence $\mathcal{T}^{base}=\{\mathcal{T}^{base}_i\}_{i=1}^p$, where $p$ denotes the length of the sequence. 
Let $\mathcal{Y}^{base}_i$ denote the label space of $\mathcal{T}^{base}_i$, we have $\bigcup_{i=1}^{p}\mathcal{Y}^{base}_i=\mathcal{Y}_0$ and $\bigcap_{i=1}^{p}\mathcal{Y}^{base}_i=\emptyset$.
In the $i$-th training iteration, let $\theta_{i-1}^{(m)}$ denote the current model parameters, which have undergone the previous $i-1$ training rounds.
Then we randomly split task $\mathcal{T}^{base}_i$ into a support set $\mathcal{S}^{base}_i$ and a query set $\mathcal{Q}^{base}_i$. 
The model first needs to perform the inner loop of meta-training on $\mathcal{S}^{base}_i$. 
Assuming the inner loop is repeated $m$ times, we have 
\begin{equation}
\theta_i^{(j)}=\theta_i^{(j-1)}-\alpha\nabla_\theta\mathcal{L}(f_{\theta_i^{(j-1)}};\mathcal{S}^{base}_i),
\end{equation}
where $0 < j \leq m$, $\theta_i^{(0)}=\theta_{i-1}^{(m)}$ and $\alpha$ denotes the inner learning rate.
After the inner loop concludes, we calculate the loss of the model $f_{\theta_i^{(m)}}$ on the query sets $\bigcup_{z=1}^{i}\mathcal{Q}^{base}_z$, denoted as $\mathcal{L}(f_{\theta_i^{(m)}};\bigcup_{z=1}^{i}\mathcal{Q}^{base}_z)$. 
We then repeat the inner loop for the next task $\mathcal{T}^{base}_{i+1}$. 
Upon completing one outer loop, we obtain the query loss for each task in the entire base task sequence. 
Next, we compute the gradient of the sum of these incremental losses (referred to as the incremental second-order gradient) with respect to the original model parameters $\theta$ and update $\theta$ as follows: 
\begin{equation}
\theta=\theta-\beta\nabla_\theta\sum_{i=1}^{p}\mathcal{L}(f_{\theta_i^{(m)}};\bigcup_{z=1}^{i}\mathcal{Q}^{base}_z),
\end{equation}
where $\beta$ denotes the outer learning rate.

The MCTF algorithm, detailed in Algorithm \ref{alg:algorithm}, facilitates this meta-learning process by aligning the model's behavior between the meta-training and incremental training, thereby enabling it to acquire prior knowledge about optimal initial parameters across a task stream and activate it during the incremental learning stage.

\begin{figure}[t]
\begin{algorithm}[H]
\renewcommand{\algorithmicrequire}{\textbf{Input:}}
\renewcommand{\algorithmicensure}{\textbf{Output:}}
\caption{An epoch of MCTF.}
\label{alg:algorithm}
\begin{algorithmic}[1]
\REQUIRE Base task $\mathcal{T}_0$; Model's parameter $\theta$.
\renewcommand{\algorithmicrequire}{\textbf{Param.:}}
\REQUIRE Inner learning rate $\alpha$; Outer learning rate $\beta$; Inner loop iterations $m$; Loss function $\mathcal{L}(\cdot;\cdot)$ that is determined by the GCL method.
\renewcommand{\algorithmicrequire}{\textbf{Input:}} 
\ENSURE Model's parameter $\theta$ that performs well across an entire stream of tasks.
\STATE Sample an ordered task sequence $\mathcal{T}^{base}=\{\mathcal{T}^{base}_i\}_{i=1}^p$ from $\mathcal{T}_0$
\STATE $\theta_0^{(m)} \gets \theta$
\FOR{$\mathcal{T}^{base}_i$ in $\mathcal{T}^{base}$}
\STATE Split task $\mathcal{T}^{base}_i$ into $\mathcal{S}^{base}_i$ and $\mathcal{Q}^{base}_i$
\STATE $\theta_i^{(0)} \gets \theta_{i-1}^{(m)}$
\FOR{$j \gets 1 $ to $m$}
\STATE $\theta_i^{(j)} \gets \theta_i^{(j-1)}-\alpha\nabla_\theta\mathcal{L}(f_{\theta_i^{(j-1)}};\mathcal{S}^{base}_i)$
\ENDFOR
\STATE Compute  $\mathcal{L}(f_{\theta_i^{(m)}};\bigcup_{z=1}^{i}\mathcal{Q}^{base}_z)$ 
\ENDFOR
\STATE $\theta\gets\theta-\beta\nabla_\theta\sum_{i=1}^{p}\mathcal{L}(f_{\theta_i^{(m)}};\bigcup_{z=1}^{i}\mathcal{Q}^{base}_z)$
\end{algorithmic}
\end{algorithm}
%\vspace{-2em}
\end{figure}

\subsection{Loss Function Design for Behavioral Alignment}

Another crucial aspect of MEGA is its approach to jointly optimizing the model across three distinct phases: the inner loop and outer loop of the meta-learning stage, as well as the incremental learning stage.
This optimization strategy emphasizes the structure of the loss function, which is essential for effectively integrating these phases.
In the context of GCL, various GCL methods typically alleviate the problem of catastrophic forgetting by modifying the loss function calculation during the incremental learning stage. 
Therefore, when incorporating GCL methods into MCTF, it is necessary to modify the loss functions in all three phases to align their training behaviors and ensure consistency.

In MEGA, we introduce the Knowledge Distillation with Single-Instance Replay (KDSIR) module as a novel GCL approach within MCTF \cite{KDER1,KDER2}. 
In the following content, we will first introduce the components of KDSIR, and then use KDSIR as an example to illustrate how to design loss functions for transferring various GCL methods to the GFSCIL scenario.

During continual learning, Knowledge Distillation (KD) uses the model trained on task $\mathcal{T}_{i-1}$ as a teacher to guide the training process on task $\mathcal{T}_{i}$. This helps ensure that the model parameters do not change excessively.
In MEGA, we optimize the model using the Mean Squared Error (MSE) loss between the outputs of the teacher model and the student model on the same data as the loss function. Formally, we have:
\begin{equation}
    \mathcal{L}_{KD}(\theta_{t},\theta_{s};x_{p})=\frac{1}{bc}(\|f_{\theta_{t}}(x_{p})-f_{\theta_{s}}(x_{p})\|_2)^2,
\end{equation}
where $f_{\theta_{t}}$ and $f_{\theta_{s}}$ represent the teacher model and the student model respectively, 
$b$ and $c$ denote the number of nodes and classes in $x_{p}$ respectively,
and $x_{p}$ denotes the input data used for training the teacher model during task $\mathcal{T}_{i-1}$. 

In GFSCIL, to reduce storage consumption and improve model generalization, we propose Single-Instance Replay (SIR), which involves randomly selecting a single sample node from the support set for each class.
Using the Cross-Entropy (CE) loss function as an example, the formal expression for the SIR loss function is as follows:
\begin{equation}
    \mathcal{L}_{SIR}(\theta;x_{b},y_{b})=\mathbf{CE}(f_\theta(x_{b}),f_\theta(y_{b})),
\end{equation}
where $x_{b}$ and $y_{b}$ denote the feature set and label set of the nodes stored in the buffer, respectively.

\noindent\textbf{Loss functions of MEGA.}
We define the complete loss functions for MEGA. Using the cross-entropy loss function as an example, the loss functions for the three phases are expressed as follows:
\begin{equation}
\begin{aligned}
    \mathcal{L}_{inner}&=\mathbf{CE}(f_{\theta_i^{(j)}}(x(\mathcal{S}^{base}_i)),y(\mathcal{S}^{base}_i))\\
    &+\mathcal{L}_{KD}(\theta_i^{(0)},\theta_i^{(j)};x(\mathcal{B}_m))\\
    &+\mathcal{L}_{SIR}(\theta_i^{(j)};x(\mathcal{B}_m),y(\mathcal{B}_m)),
\end{aligned}
\end{equation}
where $\mathcal{L}_{inner}$ represents the loss function of the inner loop in the meta-training stage. 
$\mathcal{B}_m$ is the buffer utilized during meta-training, which is cleared at the end of each epoch.
\begin{equation}
\begin{aligned}
    \mathcal{L}_{outer}&=\mathbf{CE}(f_{\theta_i^{(m)}}(x(\mathcal{Q}^{base}_i))),y(\mathcal{Q}^{base}_i))\\
    &+\mathcal{L}_{KD}(\theta_i^{(0)},\theta_i^{(m)};x(\bigcup_{j=1}^{i-1}\mathcal{Q}^{base}_j)).
\end{aligned}
\end{equation}
where $\mathcal{L}_{outer}$ denotes the loss function of the outer loop in the meta-training stage.
It is noteworthy that, since the support set is not included in the input of the outer loop of meta-training, the loss at this phase does not include $\mathcal{L}_{SIR}$.
\begin{equation}
\begin{aligned}
    \mathcal{L}_{inc}&=\mathbf{CE}(f_{\theta_i}(x(\mathcal{S}_i)),y(\mathcal{S}_i))\\
    &+\mathcal{L}_{KD}(\theta_{i-1},\theta_i;x(\mathcal{B}))\\
    &+\mathcal{L}_{SIR}(\theta_i;x(\mathcal{B}),y(\mathcal{B})),
\end{aligned}
\end{equation}
where $\mathcal{L}_{inc}$ denotes the loss function for the incremental learning stage.
$i$ represents the $i$-th task of this stage, and $\mathcal{B}$ refers to the buffer utilized during this stage.

Figure \ref{fig:framework} illustrates a schematic of a single meta-training epoch of the entire MEGA method, showing the application scenarios of the three types of loss functions mentioned above.
Importantly, MEGA can utilize not only KDSIR as the GCL module but also other established GCL methods.

\subsection{Computational Complexity of MEGA}

In this section, we elaborate on the time and space complexity analysis of MEGA.

\noindent\textbf{Time Complexity.} The meta-learning algorithm of MEGA exhibits temporal complexity equivalence to the classical Model-Agnostic Meta-Learning (MAML) algorithm. When computing second-order gradients, MEGA circumvents direct Hessian matrix computation. Instead, it approximates the second-order gradient calculation by propagating gradients back to pre-forward-propagation parameters after performing $m$ forward passes on the task sequence. This approach eliminates the quadratic time complexity inherent in Hessian computations, enabling linear-time complexity for second-order gradient calculations. For $m$ inner-loop steps and $p$ outer-loop iterations, the computational cost scales as $\mathcal{O}(p \cdot m \cdot d)$, where computational graph backtracking scales linearly with the model dimension $d$ and accumulates across $p$ tasks with $m$ inner steps each.

\noindent\textbf{Space Complexity}. MEGA's spatial complexity arises from three primary components. First, intermediate state storage during outer-loop computation requires storing parameter states$\theta_i^{(0)}$ to $\theta_i^{(m)}$ for all $p$ tasks, contributing $\mathcal{O}(p \cdot m \cdot d)$ where $d$ is the model dimension. Second, computation graph overhead for enabling second-order gradients necessitates retaining forward/backward states across $p$ tasks and $m$ inner steps, scaling as $\mathcal{O}(p \cdot m \cdot \text{activations})$ with $\text{activations}$ denoting total activation values in the graph. Additionally, the KDSIR buffer memory stores minimal class-wise single-node replays, introducing a negligible $\mathcal{O}(c)$ term ($c$ = number of classes). Collectively, the dominant spatial complexity is $\mathcal{O}(p \cdot m \cdot d)$, scaling linearly with model size $d$, task count $p$, and inner-loop adaptation steps $m$.

\section{Experiments}

\subsection{Experimental Details}

\noindent\textbf{Datasets.} 
We explore several public datasets in the GFSCIL scenarios and select the four most representative graph node classification datasets for our experiments: Amazon-Clothing \cite{amazon}, DBLP \cite{DBLP}, Cora-Full \cite{cora} and Reddit \cite{GraphSAGE}. 
These datasets feature a large number of nodes and extensive label spaces, making them ideal for evaluating the performance of GFSCIL methods. 
The statistical information for these datasets is presented in Table \ref{tab:data}. 
As a popular large-scale graph dataset containing over 200k nodes, Reddit is primarily employed to verify MEGA's scalability in high-volume network environments.
We divide each dataset into a base task and multiple novel tasks, ensuring no class overlap between tasks.

\begin{table}[t]
\caption{Statistical information of the experimental datasets.References: Amazon Clothing~\cite{amazon}, DBLP~\cite{DBLP}, Cora-Full~\cite{cora}, Reddit~\cite{GraphSAGE}.}
\centering
\footnotesize
\renewcommand{\arraystretch}{1.15}
\setlength{\tabcolsep}{3.5pt}
\begin{tabular}{@{}lcccc@{}}
\toprule
\textbf{Dataset} & 
\multicolumn{1}{c}{\textbf{Amazon Clothing}} & 
\multicolumn{1}{c}{\textbf{DBLP}} & 
\multicolumn{1}{c}{\textbf{Cora-Full}} & 
\multicolumn{1}{c}{\textbf{Reddit}} \\
\cmidrule(lr){1-5} 
\# nodes         & 24,919    & 40,672     & 19,793     & 232,965      \\ 
\# edges         & 91,680    & 288,270    & 126,842    & 114,848,857  \\ 
\# features      & 9,034     & 7,202      & 8,710      & 602          \\ 
\# labels        & 77        & 137        & 70         & 41           \\ 
\midrule
Meta-train       & 50        & 87         & 46         & 20           \\ 
Evaluation       & 27        & 50         & 24         & 21           \\ 
\bottomrule
\end{tabular}
\label{tab:data}
\end{table}

\noindent\textbf{Dataset Details.}
We evaluate our proposed MEGA on four public graph datasets as follows:
\begin{itemize}
    \item \textbf{Amazon Clothing} is a product network composed of items in the "Clothing, Shoes, and Jewelry" category on Amazon. In this dataset, each product is treated as a node, with its description serving to define the node attributes. We establish connections between products based on the substitutable relationship indicated by "also viewed".
    \item \textbf{DBLP} is a citation network of academic papers, with each node representing a paper and the links indicating the citation relationships between them. The abstracts of the papers are utilized to create the node attributes, while the class label for each node is determined by the publication venue. DBLP is widely used for research in areas like citation analysis, network analysis, and machine learning. The dataset supports various tasks, including paper recommendation and author collaboration analysis, making it a valuable resource for academic research.
    \item \textbf{Cora-Full} is a widely used graph dataset containing 19,793 scientific papers across various computer science domains. Each paper is represented as a node, with edges indicating citation relationships. The dataset includes 70 categories for classification and features are derived from a bag-of-words model with a dimensionality of 8,710. It serves as a benchmark for evaluating graph neural networks and semi-supervised learning algorithms.
    \item \textbf{Reddit} is a social network dataset derived from an online forum, containing 232,965 posts published in September 2014. Each node represents a post, and edges indicate interaction relationships where two posts are commented on by the same user, with an average node degree of 492. Node labels correspond to the community (subreddit) to which the post belongs, covering 50 active vertical topic categories. Node features are concatenated from four components: (i) the average GloVe word vectors (300-dimensional) of the post title, (ii) the average GloVe vectors of comment content, (iii) the post score (e.g., upvotes), and (iv) the total number of comments. It is commonly used for research in community detection, graph neural networks (GNNs), and dynamic graph modeling tasks.
\end{itemize}

\noindent\textbf{Baselines.} We compare our proposed method with several state-of-the-art graph few-shot learning (GFSL), graph continual learning (GCL), 
and graph few-shot class-incremental learning (GFSCIL) methods. For GFSL, we select the finetuned Graph Attention Network (GAT-FT) \cite{GAT,fine-tune}, Graph Prototype Network (GPN) \cite{GPN}, and Meta-GNN \cite{Meta-gnn}. For GCL, we select Feature-Topology Fusion-Based Experience Replay (FTF-ER) \cite{FTF-ER}, Memory Aware Synapses (MAS) \cite{MAS}, and Topology-aware Weight Preserving (TWP) \cite{TWP}. 
Notably, we adapt the existing GFSCIL methods Geometer \cite{Geometer}, HAG-Meta \cite{HAG-Meta} and Mecoin \cite{Mecoin} to our proposed setting by eliminating query set leakage in their incremental learning stages. However, this modification reveals a partial degradation in their catastrophic forgetting mitigation capabilities, thereby precluding equitable performance comparison with our method under the current setting.

\noindent\textbf{Experimental setup.} To enhance the complexity of the experiments, we configure a support set with 3 nodes and a query set with 100 nodes for each class, including those classes in the base task and streaming tasks. The number of classes for each streaming task is set to 3, 5, 3, and 3 for the Amazon Clothing, DBLP, Cora-Full, and Reddit datasets, respectively, to ensure an appropriate length of the task sequence. We use the classification accuracy of the model on the entire visible task sequence after learning the current task as the evaluation metric. To reduce fluctuation, all accuracy scores are presented as the mean and standard deviation from experiments repeated with 5 different random seeds.

\noindent\textbf{Implementation Details.}
We use Adam optimizer to optimize the models on the incremental learning stages and the outer loop of the meta-training stages, setting the initial learning rate to 0.005 and the weight decay to 5e-4. On the inner loop of the meta-training stages, we use stochastic gradient descent to optimize the models, setting the inner learning rate to 0.005. Table \ref{tab:hyperparameters} presents the complete list of hyper-parameters for the proposed MEGA method.

The regularizer hyper-parameter for MAS and TWP is always set to 10,000. And $\beta$ for TWP is set to 0.01. We set the buffer size of FTF-ER to the support set size for each class. The incremental fine-tuning step for each baseline is selected between 1 and 10 to achieve optimal performance.

\begin{table}[t]
\caption{Hyper-parameter list for MEGA.}
\centering
%\scriptsize
\renewcommand{\arraystretch}{1.1}
\resizebox{0.9\columnwidth}{!}{
\begin{tabular}{cc}
\hline
Hyper-parameter & Value  \\ \hline
optimizer & Adam               \\ 
outer learning rate & 0.005   \\ 
weight decay & 5e-4      \\ 
inner learning rate & 0.005              \\ 
inner loop step & 1 \\
GNN structure & nfeat, 32, 16, nclass    \\ 
dropout & 0.5 \\
activation & relu \\
meta-training epoch & 300 \\
incremental fine-tuning step & 5 \\ \hline
\end{tabular}}
%\vspace{-0.3cm}
\label{tab:hyperparameters}
%\vspace{-0.5cm}
\end{table}

\noindent\textbf{Environment and Code.}
We implement the MEGA method based on Pytorch framework. All evaluated models are implemented on a server with eight GPUs (NVIDIA RTX 3090 $\times$ 8).
Upon acceptance of this paper, we will release the code for the proposed MEGA method as open source.

\begin{table*}[t]
\caption{
Performance comparisons of node classification accuracy on 3 datasets under the GFSCIL setting.
MEGA's improvement is calculated relative to the best baseline.
\textcolor{red}{Red} represents the best performance, while \textcolor{blue}{Blue} represents the second-best performance.
While GAT-FT outperforms MEGA on base tasks due to extensive pre-training, its performance drops sharply during incremental training, making it unsuitable for GFSCIL scenarios. Best viewed in color.
}
\centering
%%%\vspace{-0.3cm}
\small
\renewcommand{\arraystretch}{1.3}
\newcolumntype{|}[1]{!{\color{#1}\vline}}
\resizebox{\textwidth}{!}{
\tabcolsep=1.2mm
\begin{tabular}{cccccccccc>{\columncolor{lightgray}}c|{white}|{white}>{\columncolor{lightblue}}c}
\toprule
\multirow{2}{*}{Task} & \multicolumn{11}{c}{\textbf{Amazon Clothing (3-way 3-shot GFSCIL setting)}}  \\ \cmidrule{2-12} 
            & GAT-FT        &  FTF-ER       & MAS           & TWP           & GPN           & Meta-GNN  & Geometer & HAG-Meta & Mecoin   & MEGA (Ours)  & \textbf{Impr.}     \\  
\midrule 
Base        &78.30$\pm$1.70\%	&76.38$\pm$1.52\%	&76.46$\pm$1.37\%	&77.93$\pm$1.29\%	&50.38$\pm$4.62\%	&65.41$\pm$3.89\% & 53.23$\pm$3.66\% & 59.96$\pm$3.77\% &\textcolor{blue}{80.11$\pm$1.15\%}	&\textcolor{red}{81.10$\pm$1.44\%}	&+0.99\%   \\
\midrule
Task 1   &58.79$\pm$4.46\%	&71.56$\pm$2.32\%	&53.94$\pm$3.11\%	&\textcolor{blue}{71.96$\pm$1.75\%}	&45.73$\pm$2.14\%	&53.05$\pm$3.78\%	& 48.07$\pm$3.83\% & 56.37$\pm$1.62\% &52.62$\pm$2.02\% &\textcolor{red}{74.56$\pm$1.79\%}	&+2.60\%   \\
Task 2   &9.33$\pm$6.86\%	&\textcolor{blue}{65.67$\pm$2.17\%}	&47.72$\pm$3.01\%	&58.61$\pm$3.54\%	&45.10$\pm$3.76\%	&39.63$\pm$2.35\%	& 48.05$\pm$2.21\% & 53.82$\pm$1.20\% &50.31$\pm$2.11\% &\textcolor{red}{69.83$\pm$1.83\%}	&+4.16\%   \\
Task 3   &2.26$\pm$0.80\%	&60.15$\pm$2.43\%	&44.40$\pm$2.88\%	&\textcolor{blue}{61.91$\pm$1.36\%}	&43.89$\pm$3.05\%	&27.71$\pm$4.17\%	 & 44.92$\pm$2.01\% & 52.30$\pm$1.53\% &49.54$\pm$2.55\% &\textcolor{red}{65.69$\pm$1.00\%}	&+3.78\%   \\
Task 4   &1.64$\pm$0.12\%	&\textcolor{blue}{55.43$\pm$2.12\%}	&41.98$\pm$1.80\%	&54.19$\pm$2.79\%	&42.40$\pm$4.25\%	&16.71$\pm$4.16\%	& 45.65$\pm$2.37\% & 50.48$\pm$0.94\% &46.22$\pm$1.98\% &\textcolor{red}{61.43$\pm$1.56\%}	&+6.00\%   \\
Task 5   &1.44$\pm$0.49\%	&53.64$\pm$1.70\%	&39.80$\pm$1.53\%	&\textcolor{blue}{54.42$\pm$1.52\%}	&42.45$\pm$2.95\%	&10.96$\pm$3.97\%	& 43.34$\pm$1.99\% & 50.48$\pm$1.31\% &48.67$\pm$1.62\% &\textcolor{red}{58.36$\pm$1.73\%}	&+3.94\%   \\
Task 6   &1.60$\pm$0.39\%	&51.11$\pm$1.03\%	&38.62$\pm$1.04\%	&\textcolor{blue}{51.65$\pm$1.12\%}	&41.97$\pm$2.39\%	&7.44$\pm$2.74\%	& 42.03$\pm$2.34\% & 50.25$\pm$1.34\% &49.76$\pm$2.11\% &\textcolor{red}{55.69$\pm$1.69\%}	&+4.04\%   \\
Task 7   &1.46$\pm$0.14\%	&\textcolor{blue}{49.47$\pm$1.55\%}	&37.09$\pm$1.60\%	&49.24$\pm$1.59\%	&41.45$\pm$2.74\%	&5.45$\pm$2.18\%	 & 41.99$\pm$2.01\% & 48.42$\pm$2.64\% &49.31$\pm$1.09\% &\textcolor{red}{53.07$\pm$1.70\%}	&+3.60\%   \\
Task 8   &1.36$\pm$0.03\%	&47.69$\pm$1.37\%	&36.13$\pm$1.64\%	&47.41$\pm$1.20\%	&40.54$\pm$2.49\%	&4.35$\pm$1.85\%	& 41.44$\pm$1.84\% & 47.31$\pm$1.42\% &\textcolor{blue}{49.12$\pm$1.14\%} &\textcolor{red}{50.78$\pm$1.82\%}	&+1.66\%   \\ 
Task 9   &1.29$\pm$0.02\%	&46.36$\pm$0.88\%	&34.36$\pm$1.45\%	&45.20$\pm$1.18\%	&39.30$\pm$2.89\%	&3.40$\pm$1.59\%	& 41.35$\pm$3.00\% & 45.85$\pm$1.33\% &\textcolor{blue}{48.82$\pm$1.96\%} &\textcolor{red}{49.11$\pm$1.75\%}	&+0.29\%   \\ 
\bottomrule 
\toprule
\multirow{2}{*}{Task} & \multicolumn{11}{c}{\textbf{DBLP (5-way 3-shot GFSCIL setting)}}  \\ \cmidrule{2-12} 
            & GAT-FT        &  FTF-ER       & MAS           & TWP           & GPN           & Meta-GNN   & Geometer & HAG-Meta & Mecoin   & MEGA (Ours)  & \textbf{Impr.}     \\  
\midrule
Base        &\textcolor{red}{48.08$\pm$2.07\%}	&45.24$\pm$1.64\%	&44.94$\pm$1.78\%	&46.14$\pm$1.64\%	&32.77$\pm$1.47\%	&34.85$\pm$1.63\%	& 33.27$\pm$0.65\% & 36.52$\pm$0.57\% &46.49$\pm$0.99\% &\textcolor{blue}{46.99$\pm$0.94\%}	&-1.09\%   \\
\midrule
Task 1   &13.68$\pm$4.79\%	&36.10$\pm$2.16\%	&28.09$\pm$6.26\%	&33.19$\pm$2.91\%	&29.95$\pm$0.88\%	&18.12$\pm$2.83\%	& 30.36$\pm$1.35\% & 32.79$\pm$0.87\% &\textcolor{blue}{41.70$\pm$1.02\%} &\textcolor{red}{43.53$\pm$1.88\%}	&+1.83\%   \\
Task 2   &2.21$\pm$0.96\%	&32.93$\pm$1.27\%	&26.44$\pm$4.40\%	&29.77$\pm$3.45\%	&29.42$\pm$0.73\%	&6.21$\pm$1.81\%	& 29.33$\pm$0.64\% & 32.56$\pm$0.97\% &\textcolor{blue}{37.80$\pm$1.34\%} &\textcolor{red}{41.42$\pm$1.05\%}	&+3.62\%   \\
Task 3   &1.01$\pm$0.09\%	&32.03$\pm$1.00\%	&25.93$\pm$3.61\%	&\textcolor{blue}{34.47$\pm$0.76\%}	&27.86$\pm$1.31\%	&1.56$\pm$0.55\%	& 29.05$\pm$1.09\% & 31.83$\pm$1.10\% &31.06$\pm$1.21\% &\textcolor{red}{39.77$\pm$1.07\%}	&+5.30\%   \\
Task 4   &1.08$\pm$0.28\%	&29.86$\pm$1.45\%	&25.06$\pm$3.07\%	&\textcolor{blue}{32.18$\pm$1.26\%}	&27.58$\pm$0.59\%	&0.97$\pm$0.17\%	& 27.70$\pm$1.53\% & 30.53$\pm$1.61\% &23.17$\pm$1.12\% &\textcolor{red}{38.12$\pm$1.15\%}	&+5.94\%   \\
Task 5   &0.85$\pm$0.09\%	&28.62$\pm$0.34\%	&23.77$\pm$2.72\%	&\textcolor{blue}{32.48$\pm$0.93\%}	&26.83$\pm$0.88\%	&0.85$\pm$0.08\%	& 27.00$\pm$1.06\% & 30.66$\pm$0.82\% &22.26$\pm$0.98\% &\textcolor{red}{36.62$\pm$1.36\%}	&+4.14\%   \\
Task 6   &0.89$\pm$0.05\%	&27.93$\pm$1.37\%	&22.82$\pm$2.31\%	&\textcolor{blue}{30.97$\pm$1.13\%}	&26.57$\pm$0.43\%	&0.86$\pm$0.01\%	& 26.53$\pm$0.94\% & 29.33$\pm$0.96\% &23.04$\pm$1.48\% &\textcolor{red}{34.69$\pm$0.84\%}	&+3.72\%   \\
Task 7   &0.83$\pm$0.04\%	&26.92$\pm$1.10\%	&21.50$\pm$2.56\%	&\textcolor{blue}{29.85$\pm$0.78\%}	&26.22$\pm$0.42\%	&0.81$\pm$0.03\%	& 25.72$\pm$1.16\% & 29.05$\pm$1.05\% &23.03$\pm$1.47\% &\textcolor{red}{33.31$\pm$0.86\%}	&+3.46\%   \\
Task 8   &0.82$\pm$0.08\%	&26.36$\pm$1.86\%	&20.24$\pm$1.97\%	&\textcolor{blue}{28.61$\pm$0.81\%}	&24.44$\pm$0.95\%	&0.80$\pm$0.02\%	& 25.46$\pm$0.80\% & 28.01$\pm$0.55\% &22.87$\pm$1.15\% &\textcolor{red}{31.95$\pm$0.97\%}	&+3.34\%   \\ 
Task 9   &0.78$\pm$0.04\%	&25.65$\pm$1.49\%	&18.56$\pm$2.20\%	&27.30$\pm$0.63\%	&24.70$\pm$0.45\%	&0.76$\pm$0.00\%	& 24.91$\pm$1.06\% & \textcolor{blue}{27.57$\pm$0.61\%} &22.99$\pm$1.25\% &\textcolor{red}{30.80$\pm$0.80\%}	&+3.23\%   \\ 
Task 10  &0.73$\pm$0.01\%	&25.59$\pm$1.44\%	&17.34$\pm$2.75\%	&\textcolor{blue}{26.42$\pm$0.78\%}	&24.66$\pm$0.57\%	&0.73$\pm$0.00\%	& 24.83$\pm$0.76\% & 26.14$\pm$0.71\% &22.11$\pm$0.79\% &\textcolor{red}{29.88$\pm$0.69\%}	&+3.46\%   \\ 
\bottomrule 
\toprule
\multirow{2}{*}{Task} & \multicolumn{11}{c}{\textbf{Cora-Full (3-way 3-shot GFSCIL setting)}}  \\ \cmidrule{2-12} 
            & GAT-FT        &  FTF-ER       & MAS           & TWP           & GPN           & Meta-GNN  & Geometer & HAG-Meta & Mecoin    & MEGA (Ours)  & \textbf{Impr.}     \\  
\midrule 
Base        &\textcolor{red}{72.64$\pm$0.57\%}	&28.74$\pm$5.49\%	&28.57$\pm$5.64\%	&37.24$\pm$5.71\%	&18.31$\pm$1.75\%	&20.11$\pm$3.34\%	& 28.07$\pm$3.52\% & 17.26$\pm$2.54\% &64.88$\pm$3.12\% &\textcolor{blue}{58.08$\pm$4.09\%}	&-14.56\%   \\
\midrule
Task 1   &\textcolor{blue}{45.90$\pm$6.29\%}	&33.02$\pm$5.76\%	&6.65$\pm$3.09\%	&33.99$\pm$5.53\%	&16.65$\pm$2.12\%	&14.17$\pm$4.19\%	& 23.70$\pm$3.95\% & 15.83$\pm$2.15\% &39.52$\pm$2.55\% &\textcolor{red}{52.74$\pm$3.89\%}	&+6.84\%   \\
Task 2   &15.91$\pm$2.96\%	&32.53$\pm$6.94\%	&5.81$\pm$1.84\%	&29.61$\pm$4.45\%	&16.41$\pm$2.10\%	&10.79$\pm$2.48\%	& 23.82$\pm$3.52\% & 15.23$\pm$1.80\% &\textcolor{blue}{39.79$\pm$2.68\%} &\textcolor{red}{49.82$\pm$2.99\%}	&+10.03\%   \\
Task 3   &4.80$\pm$1.75\%	&30.86$\pm$5.52\%	&6.38$\pm$2.08\%	&27.90$\pm$4.83\%	&15.11$\pm$2.16\%	&4.06$\pm$2.42\%	& 23.05$\pm$3.37\% & 14.38$\pm$2.17\% &\textcolor{blue}{39.48$\pm$1.99\%} &\textcolor{red}{46.21$\pm$2.84\%}	&+6.73\%   \\
Task 4   &2.71$\pm$0.55\%	&31.31$\pm$4.67\%	&6.71$\pm$2.40\%	&25.82$\pm$3.64\%	&14.45$\pm$0.63\%	&2.38$\pm$0.83\%	& 21.04$\pm$2.84\% & 14.33$\pm$2.56\% &\textcolor{blue}{34.61$\pm$2.23\%} &\textcolor{red}{43.48$\pm$2.83\%}	&+8.87\%   \\
Task 5   &1.82$\pm$0.14\%	&\textcolor{blue}{32.62$\pm$5.71\%}	&6.54$\pm$2.29\%	&24.56$\pm$3.88\%	&13.59$\pm$1.52\%	&1.81$\pm$0.30\%	& 21.83$\pm$2.70\% & 12.97$\pm$2.45\% &28.90$\pm$3.01\% &\textcolor{red}{40.76$\pm$2.78\%}	&+8.14\%   \\
Task 6   &1.73$\pm$0.14\%	&\textcolor{blue}{32.41$\pm$5.92\%}	&6.39$\pm$2.43\%	&23.70$\pm$3.34\%	&13.20$\pm$0.88\%	&1.67$\pm$0.02\%	& 19.24$\pm$1.79\% & 13.07$\pm$2.06\% &24.16$\pm$2.86\% &\textcolor{red}{38.98$\pm$2.47\%}	&+6.57\%   \\
Task 7   &1.60$\pm$0.00\%	&\textcolor{blue}{32.02$\pm$5.68\%}	&5.80$\pm$2.42\%	&22.39$\pm$3.68\%	&12.16$\pm$1.36\%	&1.56$\pm$0.12\%	& 18.79$\pm$1.71\% & 12.60$\pm$1.82\% &27.15$\pm$2.62\% &\textcolor{red}{36.60$\pm$2.63\%}	&+4.58\%   \\
Task 8   &1.53$\pm$0.00\%	&\textcolor{blue}{30.67$\pm$6.33\%}	&5.76$\pm$2.52\%	&21.43$\pm$3.68\%	&11.48$\pm$1.38\%	&1.46$\pm$0.14\%	& 18.67$\pm$2.95\% & 11.85$\pm$2.29\% &28.49$\pm$2.31\% &\textcolor{red}{34.16$\pm$2.14\%}	&+3.49\%   \\ 
\bottomrule 
\end{tabular}}
%%%\vspace{-0.5em}
\label{tab:main}
%\vspace{-2.5em}
\end{table*}

\begin{figure*}[t]
\centering
\includegraphics[width=0.99\textwidth]{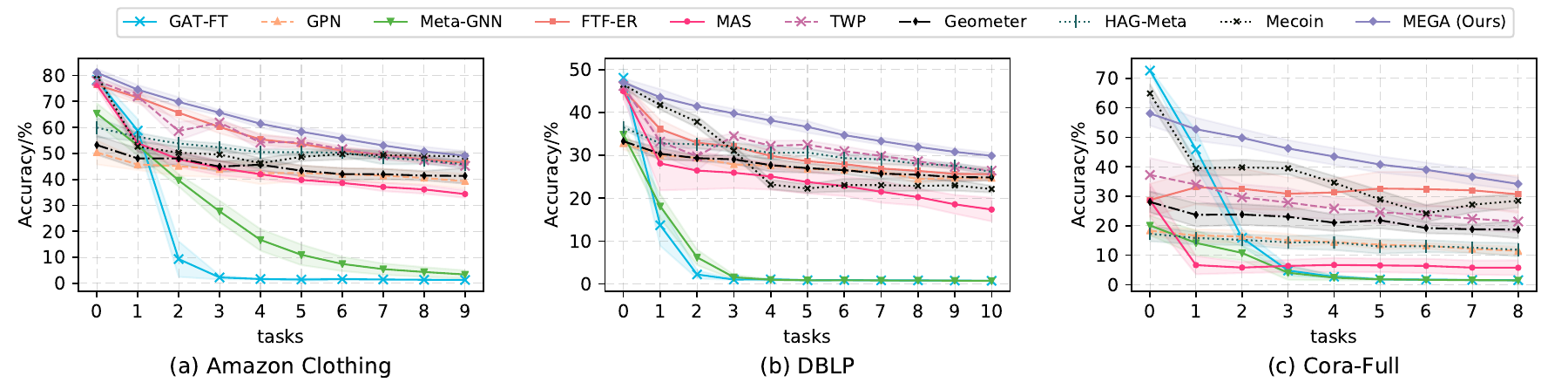} % Reduce the figure size so that it is slightly narrower than the column.
%%%\vspace{-0.3cm}
\caption{Evolution of the accuracy throughout the learning process on the task sequences of three datasets.}
\label{fig:evolution}
%%%\vspace{-1.5em}
\end{figure*}

\subsection{Comparisons with State-of-the-arts}
We compare the performance of MEGA with other baselines across three public datasets. The experimental results, presented in Table \ref{tab:main} and Figure \ref{fig:evolution}, show the average query accuracy for all encountered classes. Our comprehensive analysis indicates that MEGA consistently outperforms all baselines across the three datasets. For example, on the Cora-Full dataset, MEGA achieves performance improvements ranging from 3.49\% to 10.03\% during the incremental learning stage. 
Figure \ref{fig:evolution} visually presents the experimental results from Table \ref{tab:main}, demonstrating MEGA's superior performance.

Furthermore, compared to state-of-the-art GFSL methods (GAT-FT, GPN, Meta-GNN), MEGA demonstrates superior performance in mitigating catastrophic forgetting. 
Even though GPN and Meta-GNN exhibit lower forgetting rates, they still show a significant performance gap compared to MEGA in the final task. This further confirms MEGA's exceptional capability in GCL methods.

Additionally, compared to state-of-the-art GCL methods (FTF-ER, MAS, TWP), MEGA shows superior initial and average performance throughout the entire task sequence. The meta-training stage equips MEGA with sufficient prior knowledge, enhancing its generalization capabilities for each task. This advantage enables MEGA to outperform traditional GCL methods in few-shot scenarios.

Notably, MEGA exhibits comprehensively superior performance compared to modified GFSCIL baselines (Geometer, HAG-Meta and Mecoin) adapted to our novel experimental setting. This performance gap stems from critical architectural differences: The baseline modifications exclude incremental-stage fine-tuning while retaining only meta-learning initialization. Consequently, these methods effectively degenerate into conventional GPN during incremental learning, inheriting GPN's limited resistance to catastrophic forgetting. Empirical results demonstrate that while exhibiting mixed performance relative to GPN (-1.18\% to +10.64\%), these adapted baselines fall substantially short of MEGA's state-of-the-art performance (+3.23\% to +40.82\% gains), validating our integrated meta-incremental learning framework.

\subsection{Ablation Studies}

\begin{table}[t]
\caption{Ablation study of components on Amazon Clothing.}
\centering
%%%\vspace{0.1cm}
\small
\tabcolsep=0.7mm
\renewcommand{\arraystretch}{1.2}
\resizebox{\columnwidth}{!}{
\begin{tabular}{cccc|cccccc}
%\toprule
\hline
%\multicolumn{4}{c|}{Component} & \multicolumn{5}{c}{Acc. in each task (\%)}  \\ 
%\midrule 
%\rowcolor{lightgray} 
Setting & MCTF & $\mathcal{L}_{SIR}$ & $\mathcal{L}_{KD}$  & Base  &  1    & 3     & 5    &  7   & 9  \\  
\midrule 
\textbf{Baseline} & \ding{56} & \ding{56} & \ding{56}    & 40.1 & 41.3 & 21.6 & 7.8 & 2.5 & 1.4 \\ 
\hline
\textbf{a} & \ding{52} & \ding{56} & \ding{56}           & 78.6  & 67.2  & 34.6  & 14.3  & 7.3 & 4.6  \\ 
\textbf{b} & \ding{56} & \ding{52} & \ding{56}           & 40.3  & 46.3  & 42.4  & 36.4  & 32.2 & 28.8  \\ 
\textbf{c} & \ding{56} & \ding{56} & \ding{52}           & 45.3  & 38.6  & 23.3  & 12.4  & 5.6 & 3.2  \\ 
\hline
\textbf{d} & \ding{52} & \ding{52} & \ding{56}           & 80.2 & 72.1 & 60.8 & 55.1     &50.9  & 47.0  \\ 
\textbf{e} & \ding{52} & \ding{56} & \ding{52}           & 79.4  & 72.1  & 56.8  & 42.2  & 31.1 & 24.1  \\ 
\textbf{f} & \ding{56} & \ding{52} & \ding{52}           & 40.3  & 45.2  & 36.9  & 26.8  & 22.9 & 21.5  \\ 
\hline 
\rowcolor{lightblue} 
\textbf{g} & \ding{52} & \ding{52} & \ding{52} & \textbf{81.1} & \textbf{74.6} & \textbf{65.7} & \textbf{58.4} & \textbf{53.1} & \textbf{49.1} \\ 
\bottomrule 
\end{tabular}%
}
%\vspace{0.5em}
\label{tab:components}
%\vspace{-2.5em}
\end{table}

\begin{table}[t]
\caption{Ablation study of components on DBLP.}
\centering
\small
\tabcolsep=0.7mm
\renewcommand{\arraystretch}{1.2}
\resizebox{\columnwidth}{!}{
\begin{tabular}{cccc|cccccc}
\hline
Setting & MCTF & $\mathcal{L}_{SIR}$ & $\mathcal{L}_{KD}$  & Base  & 2    & 4     & 6    & 8   & 10  \\  
\midrule 
\textbf{Baseline} & \ding{56} & \ding{56} & \ding{56}    & 19.0 & 7.3 & 2.5 & 0.9 & 0.2 & 0.7 \\ 
\hline
\textbf{a} & \ding{52} & \ding{56} & \ding{56}           & 43.3	&	5.8&	1.8&	0.7&	0.9	&0.8 \\ 
\textbf{b} & \ding{56} & \ding{52} & \ding{56}           & 19.0	&	11.7	&	12.0	&	11.3	&	8.9&	7.7 \\ 
\textbf{c} & \ding{56} & \ding{56} & \ding{52}           & 19.0	&	6.6		&3.4	&	2.2	&	1.5	&	1.0 \\ 
\hline
\textbf{d} & \ding{52} & \ding{52} & \ding{56}           & 44.0	&	37.0	&	33.4	&	31.2	&	29.6	&	26.3 \\ 
\textbf{e} & \ding{52} & \ding{56} & \ding{52}           & 40.4	&	32.2	&	21.5	&	8.9	&10.1	&	8.3\\ 
\textbf{f} & \ding{56} & \ding{52} & \ding{52}           & 29.4	&	22.7	&	10.1&		4.9	&	6.2	&	6.5 \\ 
\hline 
\rowcolor{lightblue} 
\textbf{g} & \ding{52} & \ding{52} & \ding{52} & \textbf{47.0}	&	\textbf{41.4}	&	\textbf{38.1}	&	\textbf{34.7}	&	\textbf{32.0}	&	\textbf{29.9} \\ 
\bottomrule 
\end{tabular}%
}
\label{tab:components_dblp}
\end{table}

\begin{table}[t]
\caption{Ablation study of components on Cora-Full.}
\centering
\small
\tabcolsep=0.7mm
\renewcommand{\arraystretch}{1.1}
\resizebox{0.96\columnwidth}{!}{
\begin{tabular}{cccc|cccccc}
\hline
Setting & MCTF & $\mathcal{L}_{SIR}$ & $\mathcal{L}_{KD}$  & Base  &  2    & 4     & 6    &  8   \\  
\midrule 
\textbf{Baseline} & \ding{56} & \ding{56} & \ding{56}    & 12.3 & 4.1 & 1.8 & 1.6 & 1.5 \\ 
\hline
\textbf{a} & \ding{52} & \ding{56} & \ding{56}           & 49.5 & 21.4 & 8.1 & 3.4 & 2.1 \\ 
\textbf{b} & \ding{56} & \ding{52} & \ding{56}           & 22.3 & 5.6 & 4.9 & 4.6 & 3.5 \\ 
\textbf{c} & \ding{56} & \ding{56} & \ding{52}           & 22.3 & 3.4 & 1.8 & 1.6 & 1.5 \\ 
\hline
\textbf{d} & \ding{52} & \ding{52} & \ding{56}           & 56.2 & 44.4 & 38.1 & 36.3 & 33.4 \\ 
\textbf{e} & \ding{52} & \ding{56} & \ding{52}           & 55.1 & 48.7 & 28.5 & 14.2 & 7.1 \\ 
\textbf{f} & \ding{56} & \ding{52} & \ding{52}           & 25.0 & 14.4 & 5.8 & 3.6 & 5.7 \\ 
\hline 
\rowcolor{lightblue} 
\textbf{g} & \ding{52} & \ding{52} & \ding{52} & \textbf{58.0} & \textbf{49.8} & \textbf{43.4} & \textbf{38.9} & \textbf{34.1} \\ 
\bottomrule 
\end{tabular}%
}
\label{tab:components_cora}
\end{table}

\begin{figure}[t]
\centering
\includegraphics[width=0.99\columnwidth]{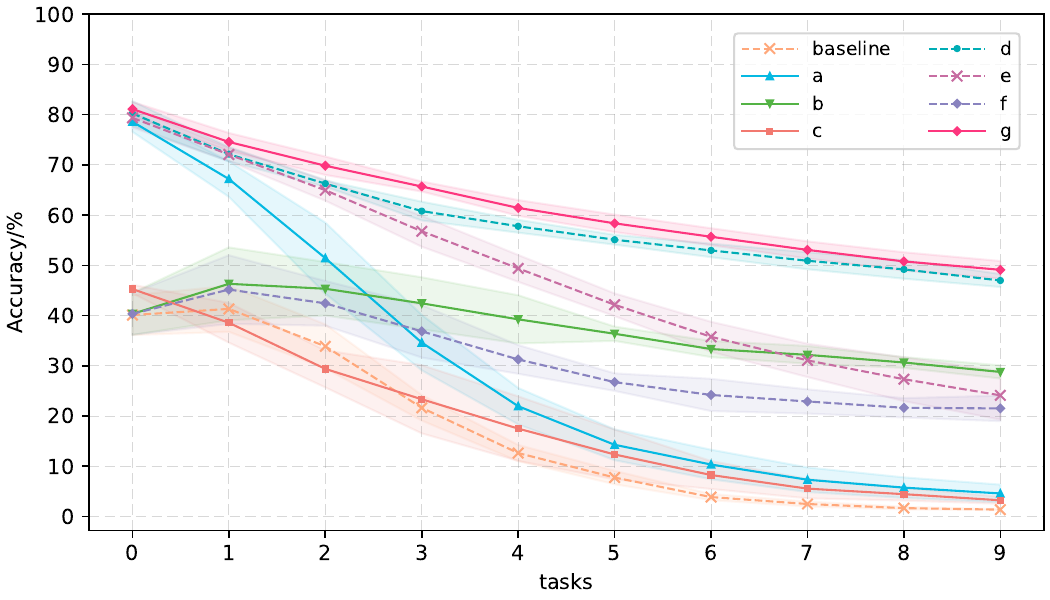} % Reduce the figure size so that it is slightly narrower than the column.
%\vspace{-0.3cm}
\caption{Evolution of the accuracy throughout the learning process on the task sequences of Amazon Clothing dataset.
Each line corresponds to a variant of MEGA in Table \ref{tab:components}.}
\label{fig:components}
%\vspace{-1.5em}
\end{figure}

\begin{figure}[t]
\centering
\includegraphics[width=0.99\columnwidth]{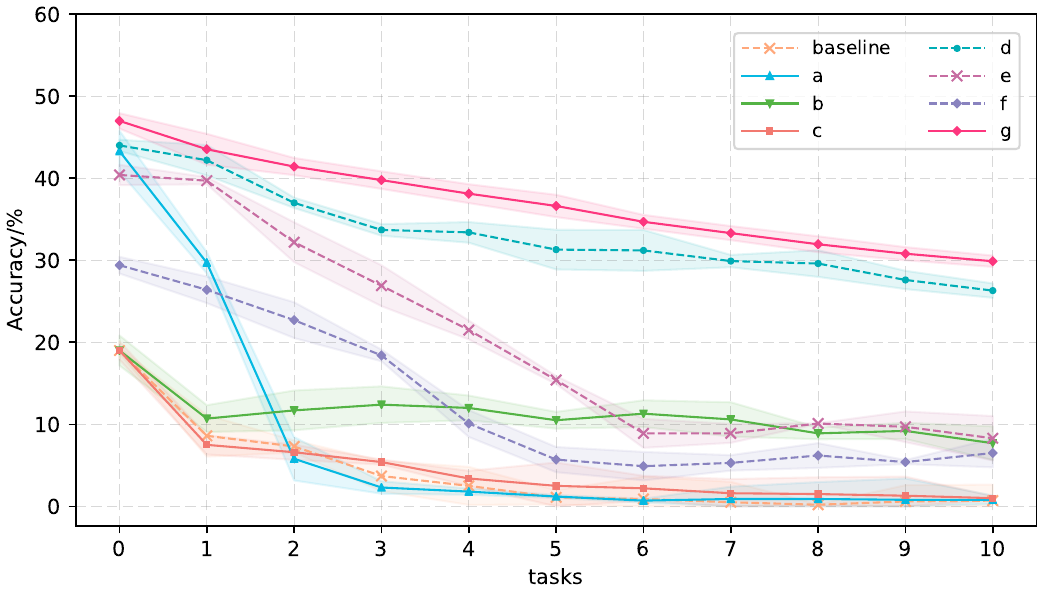} % Reduce the figure size so that it is slightly narrower than the column.
%\vspace{-0.3cm}
\caption{Evolution of the accuracy throughout the learning process on the task sequences of DBLP dataset.
Each line corresponds to a variant of MEGA in Table \ref{tab:components_dblp}.}
\label{fig:components_dblp}
%\vspace{-1.5em}
\end{figure}

\begin{figure}[t]
\centering
\includegraphics[width=0.99\columnwidth]{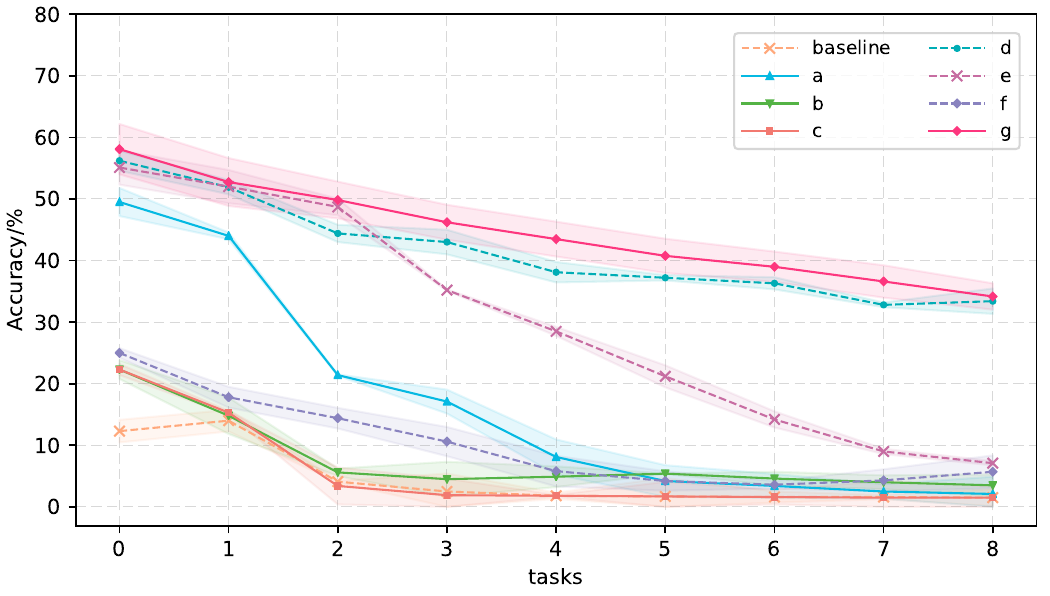} % Reduce the figure size so that it is slightly narrower than the column.
%\vspace{-0.3cm}
\caption{Evolution of the accuracy throughout the learning process on the task sequences of Cora-Full dataset.
Each line corresponds to a variant of MEGA in Table \ref{tab:components_cora}.}
\label{fig:components_cora}
%\vspace{-1.5em}
\end{figure}

\noindent\textbf{Effect of Components.} 
To validate the effectiveness of each component in our proposed MEGA method, we design a series of MEGA variants to assess the performance improvements contributed by each element. As shown in Table \ref{tab:components}, \ref{tab:components_dblp}, \ref{tab:components_cora} and Figure \ref{fig:components}, \ref{fig:components_dblp}, \ref{fig:components_cora}, removing MCTF from the model variants indicates the elimination of the entire meta-learning stage. Similarly, excluding $\mathcal{L}_{SIR}$ or $\mathcal{L}_{KD}$ signifies that these respective terms are not included in the loss function for each phase. The preceding analysis shows that incorporating a new component into a variant consistently improves model performance, confirming the significant impact of each component on overall results.

Comparing variant \textit{\textbf{a}} with \textit{\textbf{Baseline}}, \textit{\textbf{b}}, \textit{\textbf{c}}, and \textit{\textbf{f}} demonstrates that MCTF alone can raise the model's performance ceiling through the meta-learning process. However, without the integration of GCL methods, it is unable to effectively address catastrophic forgetting. Additionally, contrasting \textit{\textbf{d}}, \textit{\textbf{e}}, and \textit{\textbf{g}} with \textit{\textbf{b}}, \textit{\textbf{c}}, and \textit{\textbf{f}} reveals that incorporating GCL methods within MCTF consistently enhances the performance of these methods in few-shot scenarios. This improvement suggests that MCTF facilitates the model's ability to learn high-quality priors that support continual learning.

\begin{table*}[t]
\caption{A detailed comparison of MCTF-enhanced GCL methods against original GCL approaches on the Cora-Full dataset, presenting mean and standard deviation values across five independent experimental runs.}
\centering
%\small
%\tabcolsep=0.7mm
\renewcommand{\arraystretch}{1.4}
\resizebox{0.8\textwidth}{!}{
\begin{tabular}{r||c>{\columncolor{lightblue}}c||c>{\columncolor{lightblue}}c||c>{\columncolor{lightblue}}c}
\hline
Task & FTF-ER & \textbf{MCTF-FTFER} & MAS & \textbf{MCTF-MAS} & TWP & \textbf{MCTF-TWP} \\
\midrule 
Base & 28.74$\pm$5.49\% & \textbf{54.68$\pm$1.98\%} & 28.57$\pm$5.64\% & \textbf{36.80$\pm$4.06\%} & 37.24$\pm$5.71\% & \textbf{45.79$\pm$2.59\%} \\
Task 1 & 33.02$\pm$5.76\% & \textbf{49.03$\pm$1.33\%} & 6.65$\pm$3.09\% & \textbf{34.85$\pm$3.64\%} & 33.99$\pm$5.53\% & \textbf{39.74$\pm$2.64\%} \\
Task 2 & 32.53$\pm$6.94\% & \textbf{45.69$\pm$1.43\%} & 5.81$\pm$1.84\% & \textbf{32.95$\pm$3.53\%} & 29.61$\pm$4.45\% & \textbf{37.97$\pm$3.18\%} \\
Task 3 & 30.86$\pm$5.52\% & \textbf{42.58$\pm$1.23\%} & 6.38$\pm$2.08\% & \textbf{31.28$\pm$3.29\%} & 27.90$\pm$4.83\% & \textbf{35.00$\pm$3.02\%} \\
Task 4 & 31.31$\pm$4.67\% & \textbf{40.08$\pm$1.91\%} & 6.71$\pm$2.40\% & \textbf{29.31$\pm$3.31\%} & 25.82$\pm$3.64\% & \textbf{33.44$\pm$2.89\%} \\
Task 5 & 32.62$\pm$5.71\% & \textbf{37.71$\pm$1.53\%} & 6.54$\pm$2.29\% & \textbf{28.10$\pm$2.88\%} & 24.56$\pm$3.88\% & \textbf{31.74$\pm$3.04\%} \\
Task 6 & 32.41$\pm$5.92\% & \textbf{37.00$\pm$1.41\%} & 6.39$\pm$2.43\% & \textbf{26.50$\pm$2.77\%} & 23.70$\pm$3.34\% & \textbf{29.96$\pm$3.01\%} \\
Task 7 & 32.02$\pm$5.68\% & \textbf{35.85$\pm$0.90\%} & 5.80$\pm$2.42\% & \textbf{24.91$\pm$2.78\%} & 22.39$\pm$3.68\% & \textbf{28.28$\pm$3.10\%} \\
Task 8 & 30.67$\pm$6.33\% & \textbf{33.78$\pm$1.02\%} & 5.76$\pm$2.52\% & \textbf{23.74$\pm$2.69\%} & 21.43$\pm$3.68\% & \textbf{27.31$\pm$2.92\%} \\
\bottomrule 
\end{tabular}}
%\vspace{-0.5em}
\label{tab:versatilityMCTF}
%\vspace{-2em}
\end{table*}

\begin{figure}[t]
\centering
\includegraphics[width=0.96\columnwidth]{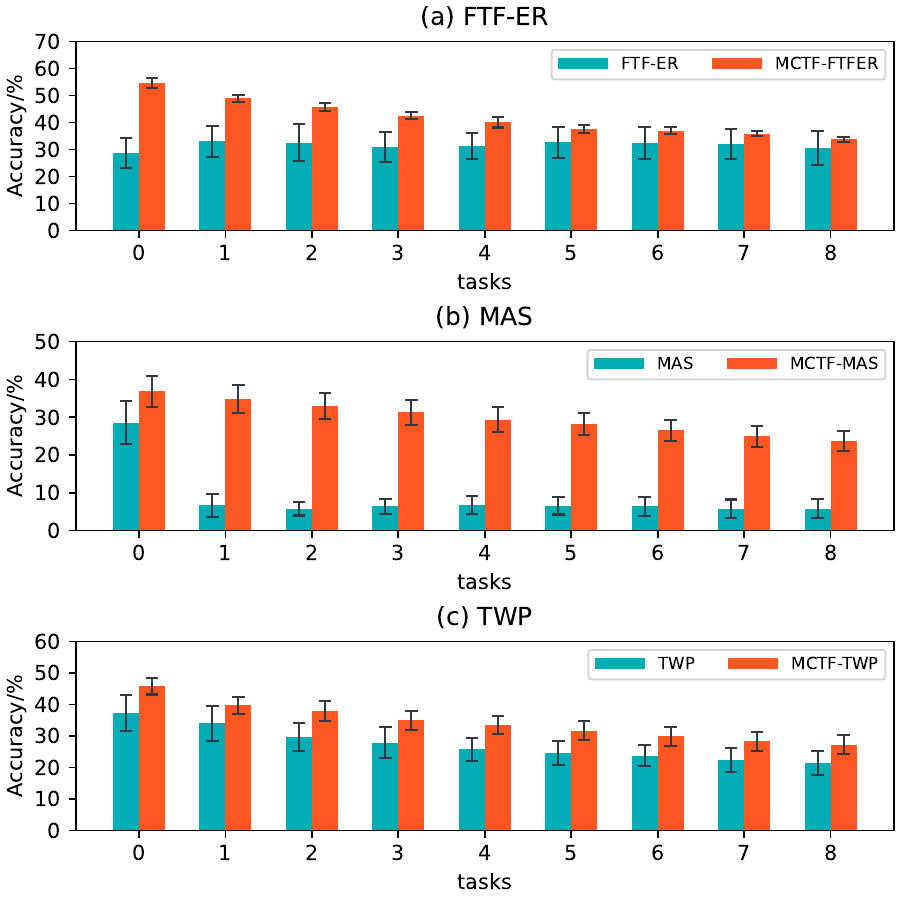} 
%\vspace{-0.2cm}
\caption{Comparison of MCTF-enhanced GCL methods against original GCL approaches on Cora-Full dataset.}
\label{fig:versatilityMCTF}
\vspace{-1em}
\end{figure}

\noindent\textbf{Versatility of MCTF.}
To demonstrate that embedding various established GCL methods into MCTF enhances their performance on GFSCIL tasks, we conduct a series of ablation experiments comparing the original GCL methods with their MCTF-embedded counterparts, focusing on FTF-ER, TWP, and MAS. The experimental results presented in Figure \ref{fig:versatilityMCTF} and Table \ref{tab:versatilityMCTF} indicate that all three MCTF-enhanced GCL methods show varying degrees of performance improvement compared to their original versions. Notably, MCTF-FTFER exhibits the most significant enhancement over FTF-ER in the early stages of incremental training, suggesting that MCTF equips FTF-ER with few-shot learning capabilities. Additionally, MCTF-MAS greatly improves MAS's ability to mitigate catastrophic forgetting in GFSCIL tasks. These findings validate the versatility of MCTF in GFSCIL settings, demonstrating that we can effectively introduce state-of-the-art GCL methods to GFSCIL tasks using MCTF.

\begin{table}[t]
\caption{Performance comparison of meta-learning techniques on DBLP dataset. MAML+CL corresponds to MCTF, while MAML+CL+KDSIR denotes our complete MEGA method.}
\centering
\small
\tabcolsep=0.7mm
\renewcommand{\arraystretch}{1.2}
\resizebox{0.96\columnwidth}{!}{
\begin{tabular}{r|ccc}
\hline
Setting & MAML & MAML+CL & \cellcolor{lightblue} \textbf{MAML+CL+KDSIR} \\ 
\midrule 
Base      & 28.85$\pm$1.67\% & 46.78$\pm$2.18\% & \cellcolor{lightblue} \textbf{46.99$\pm$0.94\%} \\ 
Task 1         & 29.59$\pm$1.63\% & 42.56$\pm$1.43\% & \cellcolor{lightblue} \textbf{43.53$\pm$1.88\%} \\ 
Task 2         & 31.01$\pm$1.72\% & 40.65$\pm$1.75\% & \cellcolor{lightblue} \textbf{41.42$\pm$1.05\%} \\ 
Task 3         & 31.01$\pm$1.43\% & 38.95$\pm$1.29\% & \cellcolor{lightblue} \textbf{39.77$\pm$1.07\%} \\ 
Task 4         & 30.71$\pm$1.08\% & 37.35$\pm$0.80\% & \cellcolor{lightblue} \textbf{38.12$\pm$1.15\%} \\ 
Task 5         & 29.71$\pm$1.26\% & 35.75$\pm$0.89\% & \cellcolor{lightblue} \textbf{36.62$\pm$1.36\%} \\ 
Task 6         & 28.50$\pm$1.22\% & 34.58$\pm$0.72\% & \cellcolor{lightblue} \textbf{34.69$\pm$0.84\%} \\ 
Task 7         & 27.28$\pm$1.41\% & 32.98$\pm$0.84\% & \cellcolor{lightblue} \textbf{33.31$\pm$0.86\%} \\ 
Task 8         & 26.37$\pm$1.24\% & 31.66$\pm$0.89\% & \cellcolor{lightblue} \textbf{31.95$\pm$0.97\%} \\ 
Task 9         & 25.61$\pm$1.05\% & 30.57$\pm$0.89\% & \cellcolor{lightblue} \textbf{30.80$\pm$0.80\%} \\ 
Task 10        & 24.44$\pm$1.02\% & 29.78$\pm$1.07\% & \cellcolor{lightblue} \textbf{29.88$\pm$0.69\%} \\ 
\bottomrule 
\end{tabular}%
}
\label{tab:effectMCTF}
\end{table}

\begin{figure}[t]
\centering
\includegraphics[width=0.95\columnwidth]{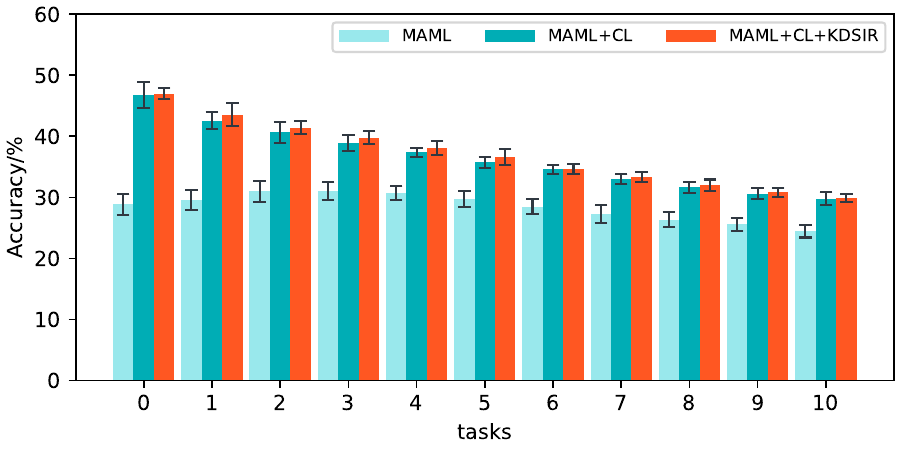} % Reduce the figure size so that it is slightly narrower than the column.
%\vspace{-0.3cm}
\caption{Comparison of various meta-learning methods on DBLP dataset. MAML+CL represents MCTF, and MAML+CL+KDSIR represents our complete MEGA method.}
\label{fig:effectMCTF}
%\vspace{-1.5em}
\end{figure}

\noindent\textbf{Effect of MCTF.} 
To further validate the rationality behind MEGA's meta-learning process, we design two variants that preserve MEGA's functionality in the incremental learning stage while altering its behavior in the meta-learning stage. In Table \ref{tab:effectMCTF} and Figure \ref{fig:effectMCTF}, \textit{\textbf{MAML}} represents a variant where MCTF is replaced by the classic MAML algorithm, \textit{\textbf{MAML+CL}} refers to a model trained using MCTF without embedded GCL modules during the meta-learning stage, and \textit{\textbf{MAML+CL+KDSIR}} denotes the complete MEGA method. The experimental results show that our proposed MCTF outperforms the original MAML algorithm, enhancing the model's adaptability to the GFSCIL setting and leading to superior performance. Additionally, incorporating KDSIR, a GCL module, into MCTF further improves the model's ability to mitigate catastrophic forgetting, thereby enhancing overall performance. These findings underscore the effectiveness of our proposed modifications to the meta-learning process and highlight the synergistic benefits of combining MCTF with GCL modules in MEGA.

\begin{figure}[t]
\centering
\includegraphics[width=0.48\textwidth]{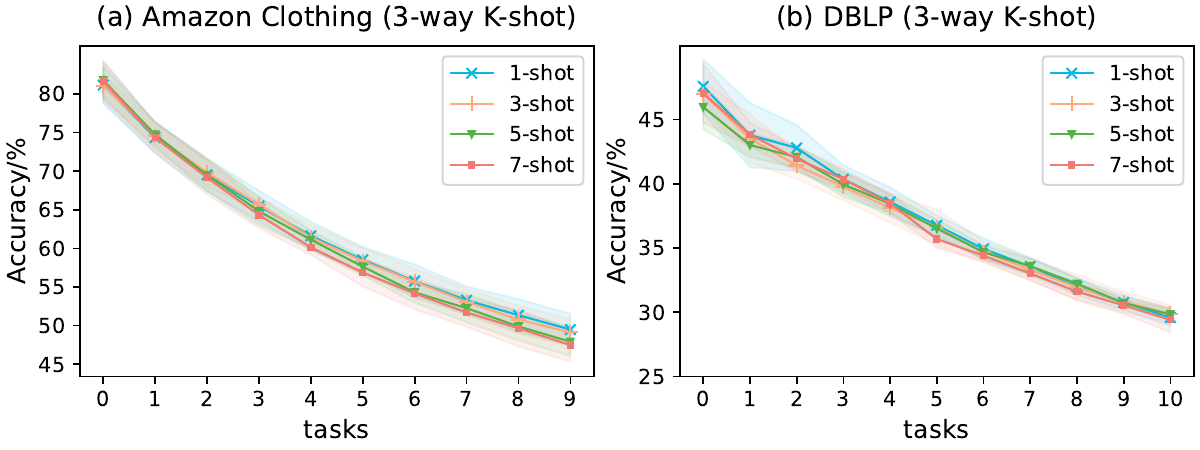} 
%\vspace{-0.8em}
\caption{The influence of support set size $K$ on Amazon Clothing and DBLP datasets.}
\label{fig:versatilityKDSIR}
%\vspace{-1em}
\end{figure}

\begin{table}[t]
\caption{Experimental results with different shot settings and their corresponding range values on the Amazon Clothing dataset (3-way).
The range, defined as the difference between maximum and minimum values, measures the effect of different shot settings on model performance across tasks.}
\centering
\small
\renewcommand{\arraystretch}{1.5}
\resizebox{\columnwidth}{!}{
\begin{tabular}{r|cccc>{\columncolor{lightblue}}c}
\hline
Task & 1-shot & 3-shot & 5-shot & 7-shot & \textbf{Range} \\
\midrule 
Base & 81.17$\pm$2.40\% & 81.10$\pm$1.44\% & 81.78$\pm$2.38\% & 81.68$\pm$2.61\% & \cellcolor{lightblue}0.68\% \\
Task 1 & 74.36$\pm$2.04\% & 74.56$\pm$1.79\% & 74.75$\pm$1.58\% & 74.35$\pm$1.95\% & \cellcolor{lightblue}0.40\% \\
Task 2 & 69.53$\pm$2.10\% & 69.83$\pm$1.83\% & 69.50$\pm$2.07\% & 69.17$\pm$2.17\% & \cellcolor{lightblue}0.66\% \\
Task 3 & 65.49$\pm$2.01\% & 65.69$\pm$1.00\% & 64.82$\pm$1.83\% & 64.33$\pm$1.69\% & \cellcolor{lightblue}1.36\% \\
Task 4 & 61.63$\pm$1.81\% & 61.43$\pm$1.56\% & 61.12$\pm$1.21\% & 60.13$\pm$0.98\% & \cellcolor{lightblue}1.50\% \\
Task 5 & 58.51$\pm$1.67\% & 58.36$\pm$1.73\% & 57.64$\pm$1.17\% & 56.87$\pm$1.79\% & \cellcolor{lightblue}1.64\% \\
Task 6 & 55.77$\pm$2.16\% & 55.69$\pm$1.69\% & 54.30$\pm$1.36\% & 54.16$\pm$2.06\% & \cellcolor{lightblue}1.61\% \\
Task 7 & 53.28$\pm$1.78\% & 53.07$\pm$1.70\% & 52.25$\pm$1.83\% & 51.70$\pm$1.86\% & \cellcolor{lightblue}1.58\% \\
Task 8 & 51.34$\pm$2.10\% & 50.78$\pm$1.82\% & 49.89$\pm$1.84\% & 49.65$\pm$2.36\% & \cellcolor{lightblue}1.69\% \\
Task 9 & 49.48$\pm$2.05\% & 49.11$\pm$1.75\% & 47.91$\pm$1.83\% & 47.48$\pm$2.13\% & \cellcolor{lightblue}2.00\% \\
\bottomrule 
\end{tabular}}
%\vspace{-0.5em}
\label{tab:versatilityKDSIR_1}
%\vspace{-2em}
\end{table}

\begin{table}[t]
\caption{Experimental results with different shot settings and their corresponding range values on the DBLP dataset (5-way).
The range, defined as the difference between maximum and minimum values, measures the effect of different shot settings on model performance across tasks.}
\centering
\small
\renewcommand{\arraystretch}{1.5}
\resizebox{\columnwidth}{!}{
\begin{tabular}{r|cccc>{\columncolor{lightblue}}c}
\hline
Task & 1-shot & 3-shot & 5-shot & 7-shot & \textbf{Range} \\
\midrule 
Base & 47.57$\pm$2.09\% & 46.99$\pm$0.94\% & 45.96$\pm$1.72\% & 47.04$\pm$2.27\% & \cellcolor{lightblue}1.61\% \\
Task 1 & 43.78$\pm$2.51\% & 43.53$\pm$1.88\% & 43.01$\pm$0.93\% & 43.83$\pm$0.93\% & \cellcolor{lightblue}0.82\% \\
Task 2 & 42.78$\pm$1.80\% & 41.42$\pm$1.05\% & 42.07$\pm$0.86\% & 41.94$\pm$0.98\% & \cellcolor{lightblue}1.36\% \\
Task 3 & 40.35$\pm$1.07\% & 39.77$\pm$1.07\% & 39.95$\pm$0.99\% & 40.35$\pm$0.78\% & \cellcolor{lightblue}0.58\% \\
Task 4 & 38.58$\pm$1.14\% & 38.12$\pm$1.15\% & 38.43$\pm$0.71\% & 38.47$\pm$0.75\% & \cellcolor{lightblue}0.46\% \\
Task 5 & 36.76$\pm$0.79\% & 36.62$\pm$1.36\% & 36.52$\pm$0.70\% & 35.69$\pm$0.67\% & \cellcolor{lightblue}1.07\% \\
Task 6 & 34.93$\pm$0.83\% & 34.69$\pm$0.84\% & 34.67$\pm$0.54\% & 34.40$\pm$0.44\% & \cellcolor{lightblue}0.53\% \\
Task 7 & 33.56$\pm$0.69\% & 33.31$\pm$0.86\% & 33.59$\pm$0.60\% & 33.01$\pm$0.50\% & \cellcolor{lightblue}0.58\% \\
Task 8 & 31.99$\pm$0.65\% & 31.95$\pm$0.97\% & 32.22$\pm$0.46\% & 31.60$\pm$0.69\% & \cellcolor{lightblue}0.62\% \\
Task 9 & 30.76$\pm$0.58\% & 30.80$\pm$0.80\% & 30.71$\pm$0.23\% & 30.53$\pm$0.64\% & \cellcolor{lightblue}0.27\% \\
Task 10 & 29.59$\pm$0.66\% & 29.88$\pm$0.69\% & 29.83$\pm$0.44\% & 29.41$\pm$0.98\% & \cellcolor{lightblue}0.47\% \\
\bottomrule 
\end{tabular}}
%\vspace{-0.5em}
\label{tab:versatilityKDSIR_2}
%\vspace{-2em}
\end{table}

\noindent\textbf{Sensitivity of Hyper-parameter.}
To investigate the impact of support set size on MEGA performance, we conduct experiments with $K$ = 1, 3, 5, and 7 on both the Amazon Clothing and DBLP datasets. Typically, in extreme scenarios such as few-shot settings, $K$ often plays a critical role in performance. However, Figure \ref{fig:versatilityKDSIR}, Table \ref{tab:versatilityKDSIR_1} and \ref{tab:versatilityKDSIR_2} show that varying $K$ has minimal influence on the results of MEGA. Notably, even in the most extreme case of $K$ = 1, our method maintains remarkably robust performance. This suggests that MEGA demonstrates excellent versatility, effectively adapting to GFSCIL tasks across various support set sizes.

\begin{table}[t]
\caption{
Performance comparisons of three GFSCIL methods on the large-scale Reddit dataset (3-way 3-shot).
}
\centering
%%%\vspace{0.2em}
\small
\tabcolsep=0.7mm
\renewcommand{\arraystretch}{1.1}
\resizebox{\columnwidth}{!}{
\begin{tabular}{r|cccccccc}
%\toprule
\hline
Method & Base & 1 & 2 & 3 & 4 & 5 & 6 & 7 \\
\midrule 
Geometer & 74.02 & 60.04 & 58.52 & 57.94 & 57.39 & 56.13 & 55.36 & 52.70\\
HAG-Meta  & 76.15 & 65.90 & 62.37 & 60.03 & 59.37 & 58.63 & 56.97 & 55.04\\ 
\rowcolor{lightblue} 
\textbf{MEGA~(Ours)}  & \textbf{90.09} & \textbf{77.45} & \textbf{70.31} & \textbf{66.57}& \textbf{64.04}& \textbf{62.85}& \textbf{60.73}& \textbf{61.10}   \\ 
\bottomrule 
\end{tabular}%
}
%\vspace{0.5em}
\label{tab:Reddit}
%\vspace{-2em}
\end{table}

\noindent\textbf{Performance on the Large-scale Dataset.} 
To validate MEGA's capability in mitigating catastrophic forgetting on large-scale datasets, comparative experiments with existing GFSCIL methods are conducted on the widely-adopted Reddit dataset. The experimental results presented in Table \ref{tab:Reddit} demonstrate that under identical settings, existing prototype-based GFSCIL approaches exhibit significantly inferior performance compared to the proposed MEGA framework. This limitation arises from two critical factors: (i) conventional GFSCIL methods inherently lose partial anti-forgetting capacity under the new GFSCIL setting, and (ii) prototype networks struggle to comprehensively capture feature representations as dataset scales increase, particularly when individual classes contain excessively large node populations. In contrast, MEGA addresses these challenges by synergistically integrating GCL mechanisms, enabling multi-scale feature extraction that fully leverages few-shot data.

\begin{table}[t]
\centering
\caption{Comparisons of memory and time overheads of various GFSCIL methods on Cora-Full dataset.}
\label{tab:cost}
\renewcommand{\arraystretch}{1.5}
\resizebox{0.99\columnwidth}{!}{
\begin{tabular}{r|ccc}
\hline
Method & Geometer\cite{Geometer} & HAG-Meta\cite{HAG-Meta} & \textbf{MEGA(Ours)} \\ \midrule 
Training Time (s) & \textbf{15.74} & 26.20 & 46.07 \\
Training Memory (MB) & 1411.31 & \textbf{1209.31} & 1397.31 \\
Average Accuracy (\%) & 22.02 & 14.17 & \textbf{44.54} \\
\hline
\end{tabular}
}
\end{table}

\noindent\textbf{Computational Overhead.} 
In Table \ref{tab:cost}, we supplement the temporal/spatial comparative experiments on the Cora-Full dataset.
Compared to existing PN-based GFSCIL methods, the incremental second-order gradient computation may result in a marginal increase in computational demands. However, as evidenced in Table \ref{tab:Reddit}, our MEGA demonstrates strong scalability on large-scale graphs, achieving accuracy improvements of +3.76\% to +17.41\% on the Reddit dataset (230K nodes). This scalability stems from two key factors:
The inherent locality of graph data and our optimized node sampling strategy.
Together, these innovations ensure robust training efficiency even as task complexity increases.

\begin{figure}[t]
\centering
\includegraphics[width=0.99\columnwidth]{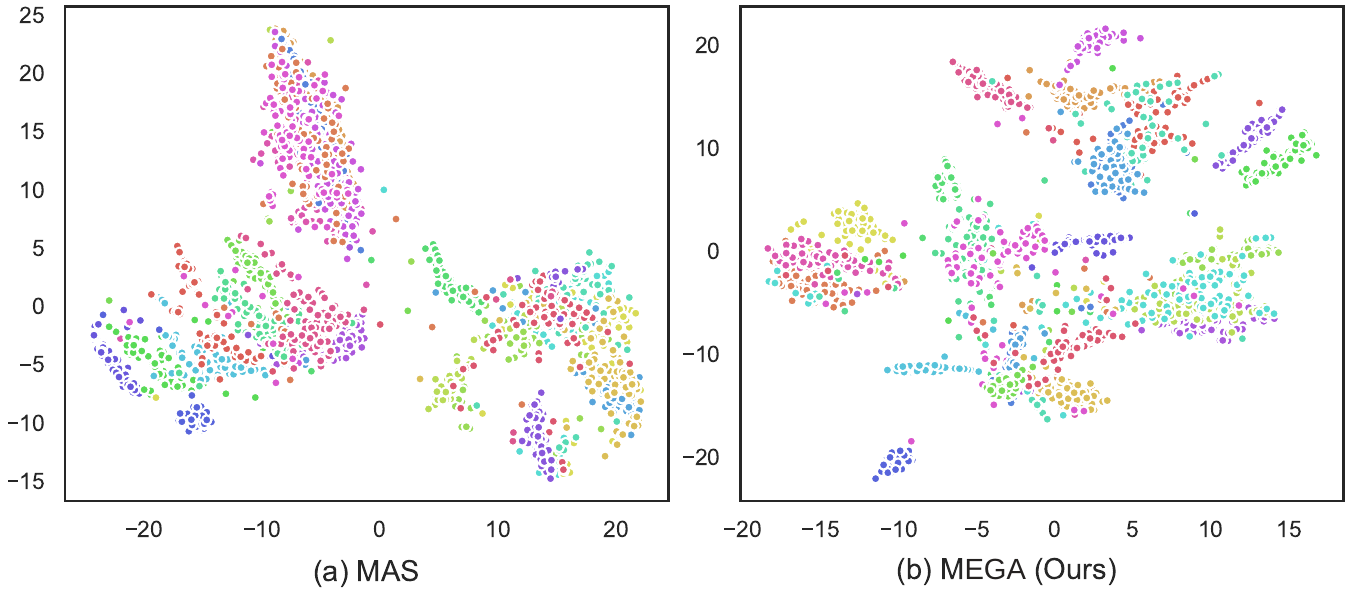} 
%\vspace{-0.5em}
\caption{2-D t-SNE projections of embeddings in two models on Amazon Clothing dataset. The nodes with different labels are represented by dots in different colors.}
\label{fig:visualization}
%\vspace{-0.5em}
\end{figure}

\noindent\textbf{Qualitative Analysis.} 
To demonstrate the effectiveness of the node representations learned by MEGA, we conduct a qualitative analysis on the Amazon Clothing dataset. For this purpose, we generate t-SNE 2D projection images \cite{t-SNE} of the hidden layer vectors from both MAS and MEGA. We select the test data from the final task of the incremental learning stage to create these hidden layer representations, which encompass 77 classes from the entire dataset. For visualization, we randomly choose one-third of these classes. The experimental results, as shown in Figure \ref{fig:visualization}, indicate that MEGA achieves a clearer separation of nodes from distinct communities compared to MAS. This observation suggests that the prior knowledge acquired by MEGA through the meta-learning process enhances the model's ability to learn novel classes while effectively mitigating catastrophic forgetting.

\section{Conclusion}

In this paper, we first propose a rigorous GFSCIL setting that restricts novel query set usage during incremental training, aligning with real-world constraints where only novel support sets are accessible for training. Building on this, we introduce MEGA (Model-Agnostic Meta Graph Continual Learning), a framework dividing training into meta-training and incremental stages. In meta-training, an incremental second-order gradient approach identifies optimal initial parameters adaptive to sequential novel tasks, while incremental training aligns behaviors with meta-training via consistent loss functions, ensuring efficient utilization of high-quality priors to mitigate catastrophic forgetting and overfitting.

Empirical validation on four prominent graph datasets confirms MEGA’s superiority, with performance improvements of 2.60\% to 7.43\% on the first novel task compared to state-of-the-art methods. By establishing a stricter GFSCIL setting and a versatile, model-agnostic paradigm compatible with existing GCL methods, MEGA not only advances GFSCIL research but also paves the way for more robust and adaptable GNN applications in dynamic, real-world scenarios.

\bibliographystyle{IEEEtran}
\bibliography{MyReference}

\end{document}